%% file: main.tex
\documentclass[10pt,twocolumn,letterpaper]{article}

\usepackage[pagenumbers]{iccv} 


\definecolor{iccvblue}{rgb}{0.21,0.49,0.74}
\usepackage[pagebackref,breaklinks,colorlinks,allcolors=iccvblue]{hyperref}

\usepackage[capitalize]{cleveref}
\usepackage{xcolor}
\usepackage{booktabs}       
\usepackage{colortbl}
\usepackage{multirow}
\usepackage{listings}
\usepackage[most]{tcolorbox}
\usepackage{tabularx}
\usepackage{graphicx}
\usepackage{subcaption}
\usepackage{makecell}
\usepackage{arydshln}
\usepackage{placeins}
\usepackage{pifont}

\newcommand{\tablestyle}[2]{\setlength{\tabcolsep}{#1}\renewcommand{\arraystretch}{#2}\centering\footnotesize}
\newlength\savewidth\newcommand\shline{\noalign{\global\savewidth\arrayrulewidth
		\global\arrayrulewidth .8pt}\hline\noalign{\global\arrayrulewidth\savewidth}}		
\newcommand\scline[1]{\noalign{\global\savewidth\arrayrulewidth
		\global\arrayrulewidth .8pt}\cline{#1}\noalign{\global\arrayrulewidth\savewidth}}

\definecolor{maroon}{cmyk}{0,0.1,0.01,0.01}
\definecolor{blue}{cmyk}{0.95,0.0,0.2,0.2}
\definecolor{yellow}{cmyk}{0.01,0.0,0.2,0.01}
\definecolor{lightblue}{cmyk}{0.1,0.0,0.02,0.02}
\definecolor{case_verb}{HTML}{fbde84}
\definecolor{case_adj}{HTML}{cccdff}
\definecolor{case_noun}{HTML}{bfeaf1}
\definecolor{case_ff}{HTML}{e65352}
\definecolor{case_error}{HTML}{ffff00}
\definecolor{darkgreen}{RGB}{51,181,41}
\definecolor{darkorange}{RGB}{252,135,62}
\definecolor{t_green}{HTML}{f1f2e4}

\definecolor{LIGHT_BLUE}{HTML}{cce4fe}
\definecolor{LIGHT_RED}{HTML}{f1b9b8}
\definecolor{LIGHT_YELLOW}{HTML}{f1f58a}
\definecolor{LIGHT_GREEN}{HTML}{f1f2e4}
\definecolor{LIGHT_PURPLE}{HTML}{b6a7b9}
\definecolor{lightgray}{gray}{0.95}

\lstdefinestyle{prompt}{
    basicstyle=\ttfamily\fontsize{7pt}{8pt}\selectfont,
    frame=none,
    breaklines=true,
    backgroundcolor=\color{lightgray},
    breakatwhitespace=true,
    breakindent=0pt,
    escapeinside={(*@}{@*)},
    numbers=none,
    numbersep=5pt,
    xleftmargin=5pt,
}
\tcbset{
  aibox/.style={
    top=10pt,
    colback=white,
    colframe=black,
    colbacktitle=black,
    enhanced,
    center,
    attach boxed title to top left={yshift=-0.1in,xshift=0.15in},
    boxed title style={boxrule=0pt,colframe=white,},
  }
}
\newtcolorbox{AIbox}[2][]{aibox, title=#2,#1}

\def\paperID{4866} 
\def\confName{ICCV}
\def\confYear{2025}

\title{Creation-MMBench: Assessing Context-Aware Creative Intelligence in MLLMs}

\author{\bf Xinyu Fang$^{1,2}$\footnotemark[1], Zhijian Chen$^{3}$\footnotemark[1], Kai Lan$^{3}$, Lixin Ma$^{3}$, \\
\bf Shengyuan Ding$^{2,4}$, Yingji Liang$^{5}$, Xiangyu Zhao$^{2,6}$, Farong Wen$^{6}$, \\ \bf Zicheng Zhang$^{2,6}$, 
Guofeng Zhang$^{1}$, Haodong Duan$^{2}$\footnotemark[2], Kai Chen$^{2}$\footnotemark[2], Dahua Lin$^{2,7}$
\\
Zhejiang University$^1$ \quad
Shanghai AI Laboratory$^2$ \quad
Tongji University$^3$  \quad
Nanjing University$^4$ \\
East China Normal University$^5$ \quad
Shanghai Jiaotong University$^6$ \quad
The Chinese University of Hong Kong$^7$
}

\begin{document}

\twocolumn[{%
\renewcommand\twocolumn[1][]{#1}%
\maketitle

\begin{center}
    \centering
    \captionsetup{type=figure}
    \vspace{-7mm}
    \includegraphics[width=\linewidth]{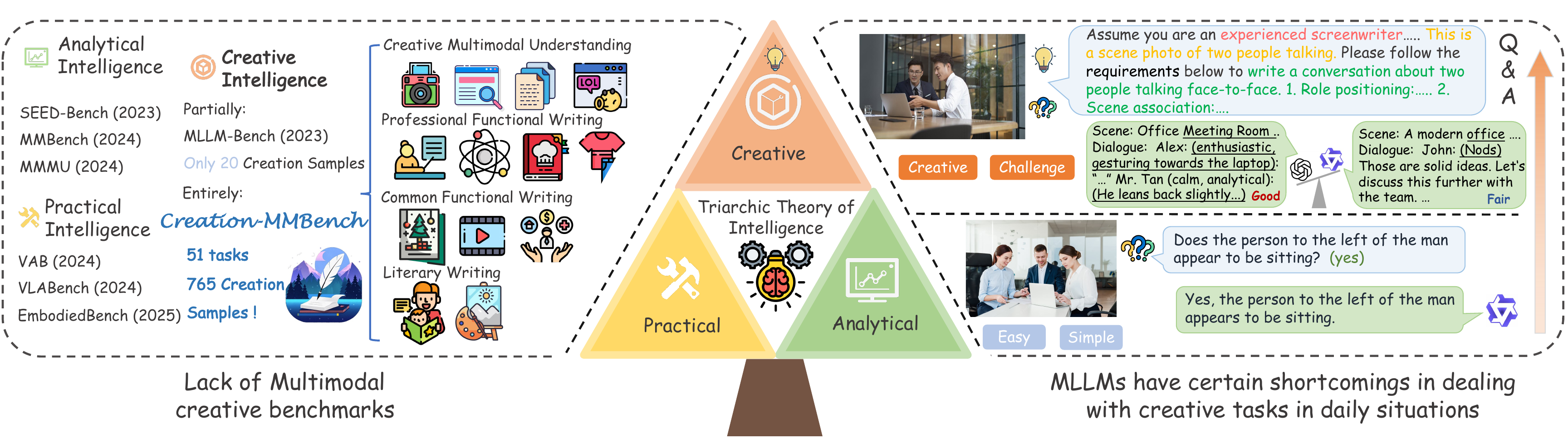}
    \vspace{-3mm}
    \caption{\textbf{Our Motivation for Creation-MMBench.} The triarchic theory of intelligence divides intelligence into three forms. Current MLLM benchmarks have significant gaps in evaluating visual-creative intelligence compared to the other forms. Additionally, existing benchmarks feature simple questions that fail to assess model performance in real-life creative tasks. Therefore, we proposed Creation-MMBench, which includes four categories, more creative and discriminative questions, and better evaluation of visual creative intelligence.}
    \label{fig:spotlight}

\end{center}%
}
]
\footnotetext[1]{Equal Contribution.}
\footnotetext[2]{Corresponding Author.}
\input{sec/0_abstract}    
\input{sec/1_intro}
\input{sec/2_related_work}

\input{sec/3_method}

\input{sec/4_experiment}

\input{sec/6_conclusion}
\newpage
{
    \small
    \bibliographystyle{ieeenat_fullname}
    \bibliography{main}
}
\input{sec/X_suppl}

\end{document}

%% file: sec/0_abstract.tex
\begin{abstract}

Creativity is a fundamental aspect of intelligence, 
involving the ability to generate novel and appropriate solutions across diverse contexts. 
While Large Language Models (LLMs) have been extensively evaluated for their creative capabilities, 
the assessment of Multimodal Large Language Models (MLLMs) in this domain remains largely unexplored. 
To address this gap, we introduce Creation-MMBench, 
a multimodal benchmark specifically designed to evaluate the creative capabilities of MLLMs in real-world, image-based tasks. 
The benchmark comprises 765 test cases spanning 51 fine-grained tasks.
To ensure rigorous evaluation, we define instance-specific evaluation criteria for each test case, 
guiding the assessment of both general response quality and factual consistency with visual inputs. 
Experimental results reveal that current open-source MLLMs significantly underperform compared to proprietary models in creative tasks. 
Furthermore, our analysis demonstrates that visual fine-tuning can negatively impact the base LLM’s creative abilities.
Creation-MMBench provides valuable insights for advancing MLLM creativity and establishes a foundation for future improvements in multimodal generative intelligence. Full data and evaluation code is released on \url{https://github.com/open-compass/Creation-MMBench}. 

\end{abstract}

%% file: sec/1_intro.tex
\section{Introduction}
\label{sec:intro}

Creativity is the ability to generate \textbf{novel} and \textbf{appropriate} solutions to complex problems across various contexts\cite{amabile2018creativity, mayer1999fifty}. 
With the rapid advancement of Large Language Models (LLMs), 
numerous benchmarks have been proposed to assess their capabilities across different dimensions of intelligence, 
including comprehension, reasoning, and creativity~\cite{wang2024mmlupro, ruan2024liveideabench,rein2024gpqa,hendrycks2021measuring,2025creativewriting}. 
These benchmarks have significantly contributed to a deeper understanding of LLM intelligence and have played a crucial role in driving their improvement.
Meanwhile, Multimodal Large Language Models (MLLMs)~\cite{liu2023visual,bai2023qwen,chen2023internvl} have also benefited from advancements in LLMs, 
achieving notable progress in perception, reasoning, and other cognitive abilities~\cite{chen2024we,yue2024mmmu,lu2023mathvista}.

\begin{figure}[t]
\centering
\includegraphics[width=0.8\linewidth]{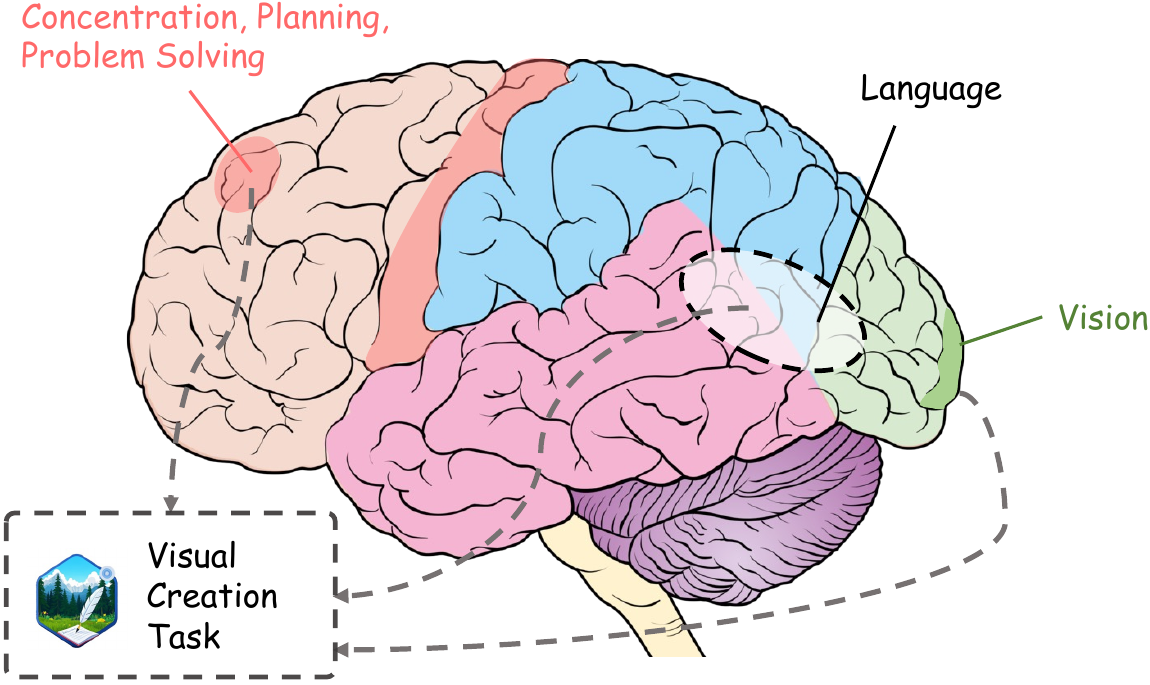}
\vspace{-1mm}
\caption{\textbf{Brain regions related to creativity and their respective functions~\cite{heilman2016possible, gao2021subcortical}.} }
\vspace{-4mm}
\label{fig:brain}
\end{figure}

As a well-established theory in psychology, 
the Triarchic Theory of Intelligence~\cite{sternberg1997triarchic} comprises three subtheories: 
the analytical subtheory, the contextual subtheory, and the creative subtheory. 
The analytical subtheory primarily focuses on information processing and problem-solving skills based on 
domain-specific knowledge and can be assessed through various knowledge and reasoning benchmarks~\cite{yue2024mmmu,hao2025can}. 
The contextual subtheory, on the other hand, 
emphasizes practical intelligence in real-world scenarios and is typically evaluated using agent-based or embodied AI benchmarks~\cite{yang2025embodiedbench,zhang2024vlabench}. 
Despite the significance of the creative subtheory in intelligence, 
evaluations of MLLMs' creative capabilities remain highly inadequate and lag significantly behind those conducted for LLMs~\cite{2025creativewriting,guo2024ideabench}.
Moreover, constructing benchmarks to assess visual creativity presents inherent challenges. 
Cognitive science research suggests that creativity arises from a distributed cortical network involving the coordination of multiple brain regions. 
As illustrated in \cref{fig:brain}, creativity is closely associated with functions of the frontal lobe, such as concentration, planning, and problem-solving~\cite{heilman2016possible}. 
Within the context of MLLM evaluation, assessing creative capabilities requires benchmarks that encompass a broader range of fundamental cognitive abilities compared to those needed for other types of intelligence assessment~\cite{liu2024mmbench,yu2023mmvet}.

To address this significant gap, 
we introduce \textbf{Creation-MMBench}, 
a novel benchmark specifically designed to assess the creative capabilities of MLLMs in image-based tasks across authentic real-world scenarios. 
The benchmark consists of 765 test cases spanning 51 fine-grained tasks,
which are categorized into four major groups: 
\textbf{Literary Writing}, \textbf{Common Functional Writing}, \textbf{Professional Functional Writing}, and \textbf{Creative Multimodal Understanding}. 
Additionally, the benchmark is accompanied by rich context to facilitate comprehensive evaluation.
In each task, an MLLM is provided with one or more images along with 
a detailed context specifying the assigned role, necessary background information, and clear task instructions. 
The model then follow the instruction and leverage the visual input to accomplish various creative tasks, 
such as composing artwork-inspired prose, 
developing structured lesson plans, 
or interpreting the conceptual foundations of advertisements. 
The approach enables a systematic assessment of MLLMs' capacity to integrate visual perception with creative expression in contextually appropriate ways.

Unlike ground-truth based evaluations, creative responses generated by models resist rule-based assessment methods. 
In our evaluation framework, 
we implement the widely adopted MLLM-as-a-Judge methodology, 
utilizing GPT-4o to assess the quality of model-generated responses. 
Given the diverse task types and stylistic variations across Creation-MMBench, 
a single-criterion evaluation model cannot reliably assess all tasks. 
To this end, we define instance-level evaluation criteria for each test case, 
ensuring that responses are assessed based on their ability to integrate contextual and visual information effectively. 
Using these tailored criteria, an MLLM-generated response is compared against a reference answer, and preferences are assigned accordingly.
In addition to the preference obtained through pairwise comparison, 
we introduce a visual factuality score to evaluate whether the MLLM's response aligns with key facts present in the visual input. 
This factual score is determined through unitary evaluation conducted by the GPT-4o judge model. 
Both Unitary Scoring and Pairwise Comparison offer a comprehensive assessment of creative quality and factual accuracy.

Based on Creation-MMBench, we conduct a comprehensive evaluation of mainstream MLLMs. 
The results indicate that current open-source MLLMs generally underperform compared to advanced proprietary models (\textit{e.g.}, Gemini-2.0-Pro, GPT-4o) in terms of context-aware creativity.
To further explore the impact of visual instruction tuning, 
we transformed Creation-MMBench into a text-only variant, 
\textbf{Creation-MMBench-TO}, 
by replacing image inputs with corresponding textual descriptions. 
The results reveal a negative effect of visual fine-tuning on the creative abilities of the base LLM, 
suggesting potential trade-offs introduced by multimodal adaptation.

In summary, our main contributions are three-fold:

\begin{enumerate}[label={\bf {{$\bullet$}}},,leftmargin=*,topsep=0.5ex,itemsep=-0.5ex,partopsep=0.75ex,parsep=0.75ex,partopsep=0pt,wide,labelindent=0pt]
    \item Development of Creation-MMBench, a multimodal benchmark specifically designed to evaluate the creative capabilities of MLLMs. The benchmark incorporates a diverse set of image sources, spans a wide range of topics and task types across real-world scenarios, and features high-quality, original human-written instructions.
    \item Design of a robust evaluation methodology that includes 
    carefully crafted instance-specific criteria for each test case, 
    enabling assessment of both general response quality and visual-factual alignment in model-generated content.
    \item A comprehensive assessment of various MLLMs on Creation-MMBench, providing detailed insights into their performance. The results highlight the current limitations of MLLMs in context-aware creativity and vision-based language generation, offering valuable guidance for future research and development.
\end{enumerate}

%% file: sec/2_related_work.tex
\begin{figure*}[t]
\centering
\includegraphics[width=0.9\linewidth]{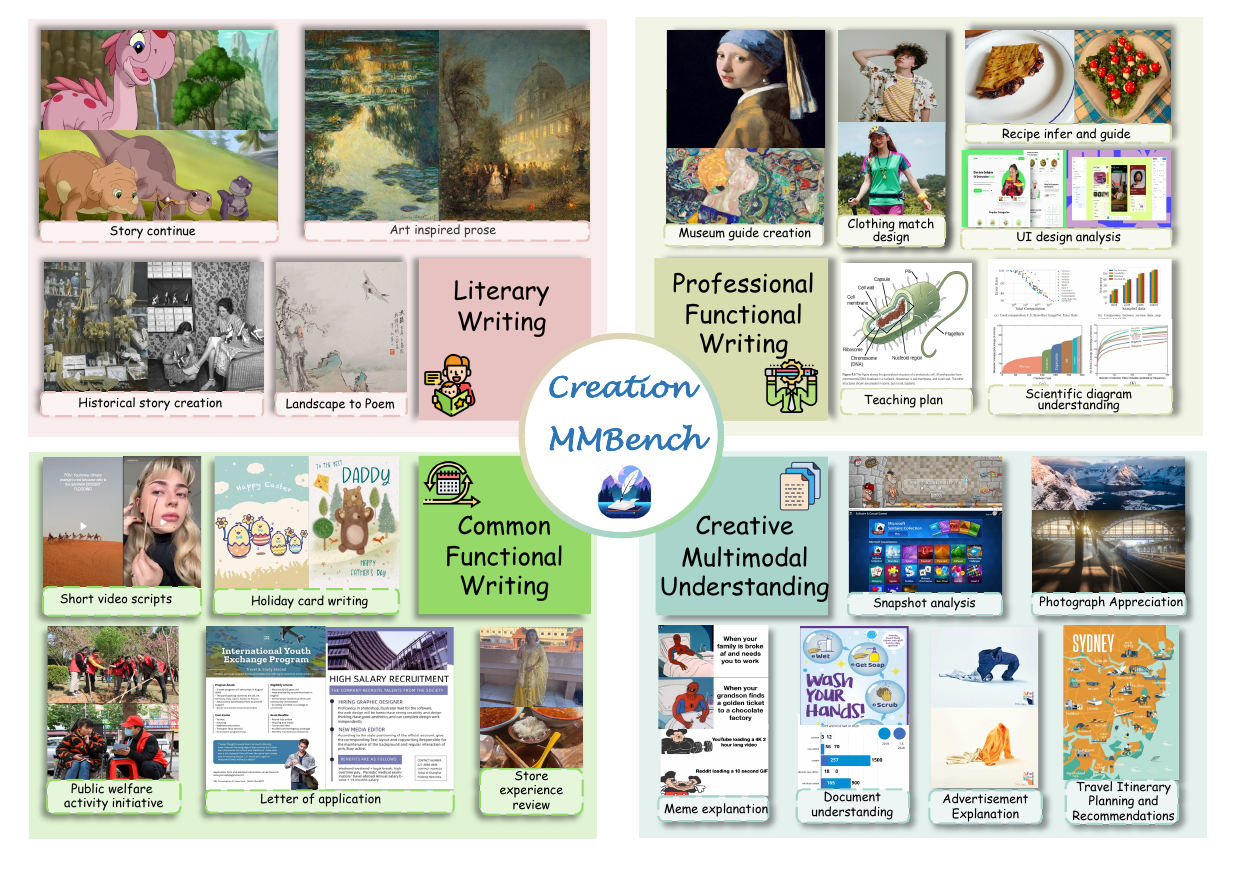}
\vspace{-3mm}
\caption{\textbf{Overview of Creation-MMBench.} Contains four task categories, each category consists of multiple tasks, and the types of images are diverse. Only a few representative tasks of each category are shown here. Complete list of tasks is detailed in the Appendix A. }
\label{fig:task-overview}
\end{figure*}

\section{Related Work}
\label{sec:related_work}

\noindent\textbf{Evaluating Creative Capabilities of LLMs.}
To evaluate the creative writing capabilities of large language models (LLMs), several benchmark tests have been introduced. One example is the LLM Creative Story-Writing Benchmark~\cite{2025creativewriting}, where 26 LLMs generate 500 short stories each, incorporating random elements, for a total of 13,000 stories. Six models then assess these stories based on 16 criteria related to character development, plot, and narrative structure. Another test ~\cite{williams2024confederacy} challenges models and humans to create stories based on specific prompts. These benchmarks assess not only the writing quality but also the diversity and complexity of the generated content.

In addition to creative writing tasks, psychological tests commonly used to assess human creativity have also been adapted for evaluating LLMs. The Alternative Uses Test (AUT) evaluates a model’s ability to propose novel uses for everyday items within a time limit, as demonstrated in the assessment of GPT-3’s creativity ~\cite{stevenson2022putting}. Another benchmark introduces a small-scale test with a leaderboard to evaluate how four LLMs generate alternative uses for objects~\cite{rabeyah2024llms}. The Torrance Tests of Creative Thinking (TTCT) have also been applied to LLMs to assess fluency, flexibility, originality, and elaboration in creative tasks~\cite{guzik2023originality}. 

Brainstorming techniques, commonly used to boost creativity, have been applied to evaluate LLMs’ creative abilities. RPGBench \cite{yu2025rpgbench} uses role-playing games to assess creativity, and LiveIdeaBench \cite{ruan2024liveideabench} evaluates scientific creativity using single-keyword prompts, focusing on novelty, feasibility, fluency, and flexibility. Other benchmarks like LLM-Evolve \cite{you2024llm} test problem-solving and adaptability, while SimulBench \cite{jia2024simulbench} evaluates creative simulations like acting as a Linux terminal. These benchmarks offer a comprehensive evaluation of LLMs' creative and simulation capabilities, inspiring further exploration of MLLMs' creative potential.

\noindent\textbf{Advancing the Evaluation of Creative Intelligence in MLLMs.}
The advancement of MLLMs has led to the development of various benchmarks to evaluate their intelligence. MMBench~\cite{liu2024mmbench} covers 20 distinct ability dimensions, focusing on MLLMs' general capability. MMMU~\cite{yue2024mmmu} evaluates advanced perception and reasoning with domain-specific knowledge, featuring 11,500 multimodal questions across 6 disciplines. These benchmarks mainly focus on the analytical intelligence of MLLMs. For assessing MLLMs' contextual intelligence, agent-based or embodied AI benchmarks are commonly used. VLABench~\cite{zhang2024vlabench} provides 100 categories of tasks to evaluate robotics' language-conditioned manipulation ability, while EmbodiedBench~\cite{yang2025embodiedbench} offers a comprehensive evaluation on models' problem-solving ability with 1,128 tasks across 4 environments.

While the evaluation of MLLMs' analytical and contextual intelligence has become relatively mature, the assessment of their creative intelligence remains insufficient. Existing partial-creation benchmarks, such as MLLM-Bench~\cite{ge2023mllm_bench} and AlignMMBench~\cite{wu2024alignmmbench}, lack a systematic and comprehensive evaluation, often failing to assess models' capabilities in complex, real-world scenarios. Furthermore, a dedicated benchmark designed specifically to evaluate MLLMs' creativity has yet to be developed. Therefore, there is a pressing need for a comprehensive and practical benchmark to bridge this gap. Creation-MMBench aims to establish a dedicated benchmark for creative ability evaluation by incorporating a diverse set of real-world tasks, offering a novel perspective on evaluating MLLMs’ creative intelligence.

\input{tables/data_basic_statistics_with_other_creation}

%% file: tables/data_basic_statistics_with_other_creation.tex
\begin{table}[t]
\centering
\resizebox{\linewidth}{!}{%
\tablestyle{1pt}{1.2}
\begin{tabular}{lccccc}
    \toprule
        \textbf{Benchmarks} & 
        \makecell{\textbf{Num of} \\ 
        \textbf{Creative} \\ \textbf{Questions}} &
        \makecell{\textbf{Criteria} \\ \textbf{Level}} &
        \makecell{\textbf{multi-images} \\ \textbf{task}} & \makecell{\textbf{Specific Role} \\ \textbf{for each} \\ \textbf{Questions}} &
                    \makecell{\textbf{Visual} \\ \textbf{Factuality} \\ \textbf{Check}}  \\
        \midrule
        VisIT-Bench & 65 & benchmark & \ding{51} & \ding{55} &  \ding{51} \\ 
        MLLM-Bench & 20 & instance & \ding{55} & \ding{55} &  \ding{51} \\
        Touch-Stone & 189 &  benchmark & \ding{51} & \ding{55} & \ding{55} \\
        AlignMMbench & 353 & task & \ding{55} & \ding{55} &  \ding{55} \\
        \midrule
        \textbf{Creation-MMBench} & \textbf{765} & \textbf{instance} &  \ding{51} & \ding{51} & \ding{51} \\
        \bottomrule
    \end{tabular}
}

\caption{\textbf{Comparison of Creation-MMBench with other partial-creation MLLM benchmarks.} 
}
\label{tab: comparsion_with_partial_creation}
\end{table}

%% file: sec/3_method.tex
\section{Creation-MMBench}
\label{method}

This section describes the construction process of Creation-MMBench, covering aspects such as task design, data collection, annotation, quality control, and evaluation. As shown in \cref{fig:task-overview}, the dataset includes diverse categories, reflecting the complexity and breadth of the tasks involved. Additionally, we introduce the data format and the indicators used to assess model capabilities. 

\subsection{Benchmark construction}

\textbf{Task Design. } 
We began with a brainstorming session to explore creative tasks in daily scenarios and designed a prototype task set encompassing both routine (e.g., writing common emails) and professional tasks (e.g., designing teaching plans). Leveraging a large language model, we then expanded this set to generate a diverse range of candidate tasks. Finally, through manual refinement and integration, a well-defined set of 51 tasks was established.

\begin{figure}[t]
\centering
\includegraphics[width=.8\linewidth]{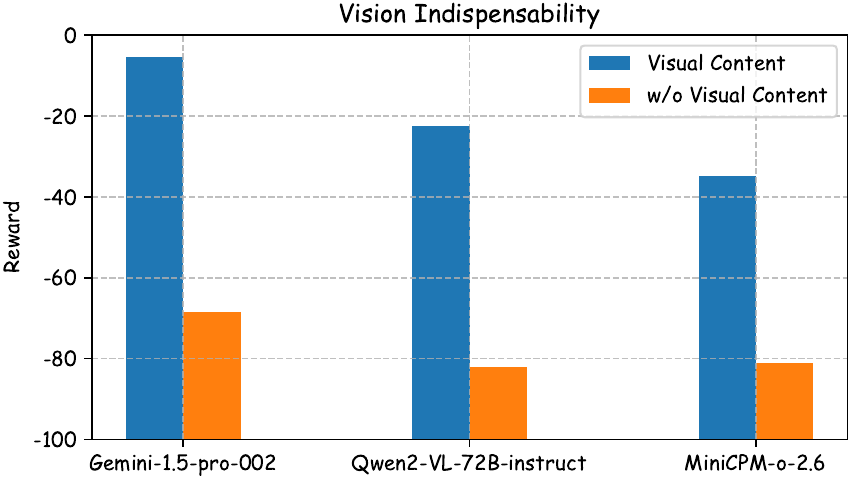}
\vspace{-1mm}
\caption{\textbf{Evaluation Result of MLLMs w/o visual input.}}
\vspace{-4mm}
\label{fig:vision_indispensability}
\end{figure}

\noindent\textbf{Task Categorization. } 
We divided the 51 tasks into four main categories: 
\begin{enumerate}
\item Literary Writing: Focus on literary creation (poetry, dialogues, stories, etc.)
\item Common Functional Writing: Focus on functional writing in daily life (social media writing, daily affairs inquiry, etc.)
\item Professional Functional Writing: Focus on functional writing and creative problem-solving in professional domains (analyzing design, developing lesson plans, etc.)
\item Creative Multimodal Understanding: Focus on the integration of visual understanding and creativity (formatted visual content analysis, image appreciation, etc.)
\end{enumerate}

\begin{figure*}[htbp]
    \centering
    \begin{subfigure}{0.27\textwidth}  
        \centering
        \begin{subfigure}{\textwidth}  
            \includegraphics[width=\linewidth]{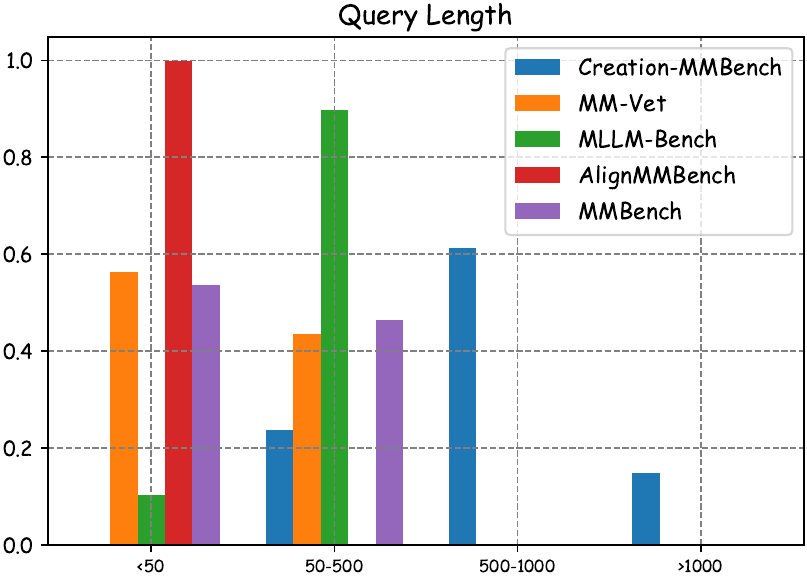}
            \caption{Distribution of query lengths.}
            \label{fig:prompt_len}
        \end{subfigure}
        \begin{subfigure}{\textwidth}  
            \includegraphics[width=\linewidth]{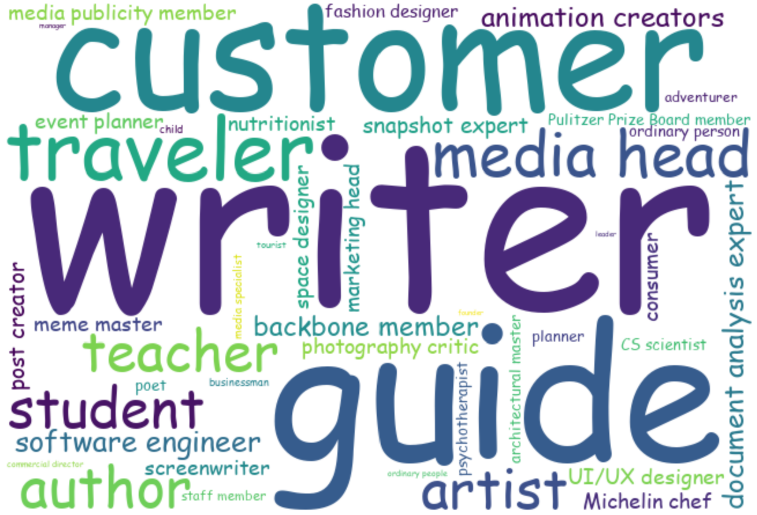}
            \caption{Roles in Creation-MMBench.}
            \label{fig:role_cloud}
        \end{subfigure}
    \end{subfigure}
    \begin{subfigure}{0.705\textwidth}  
        \centering
        \includegraphics[width=\linewidth]{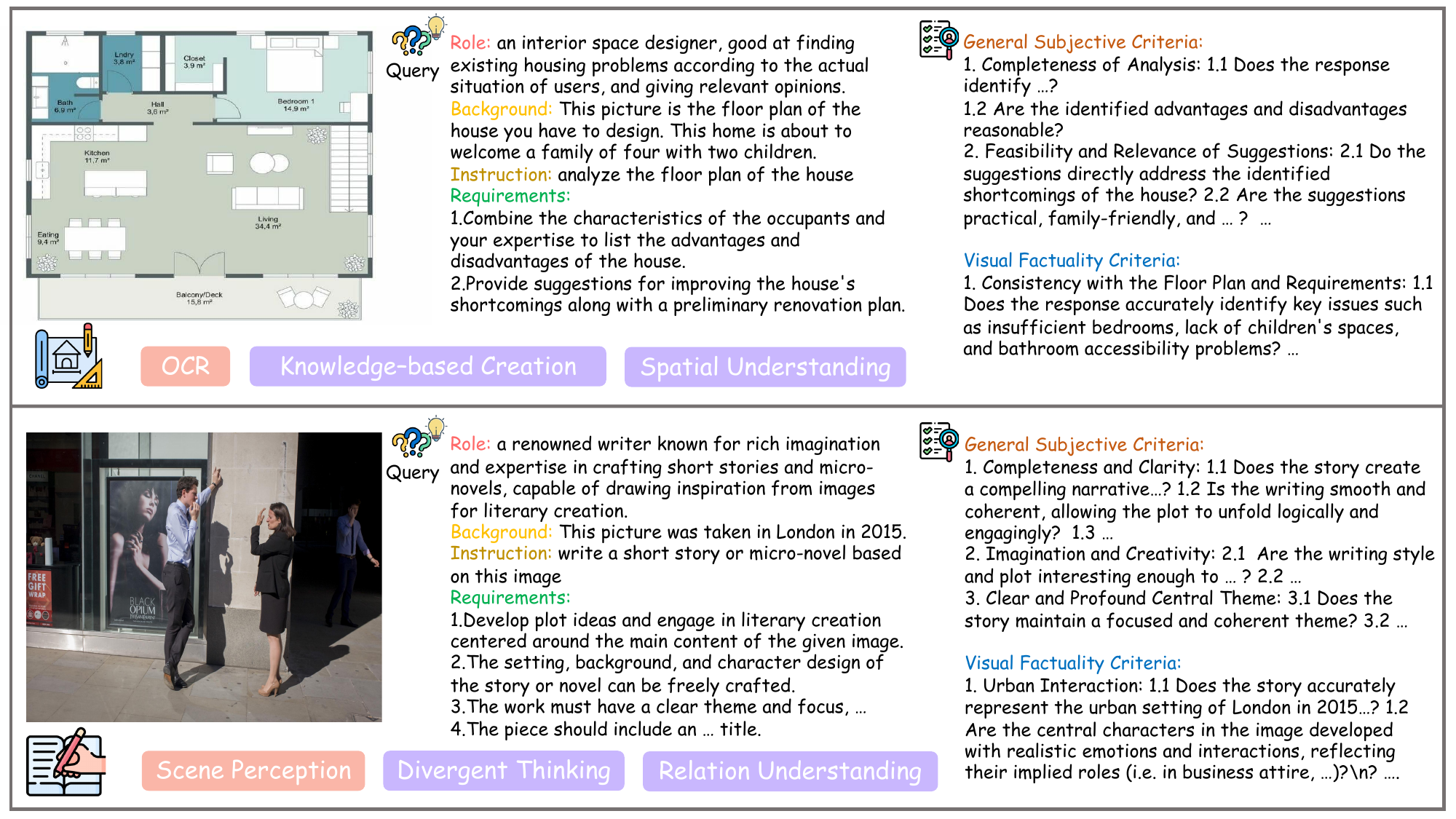}
        \caption{Example Case of Creation-MMBench.}
        \label{fig:example_case}
    \end{subfigure}
    \caption{\textbf{Statistics and Cases of Creation-MMBench.} Compared to other widely used MLLM benchmarks, Creation-MMBench features a more comprehensive query design to capture abundant creative contexts. Diverse roles are introduced into the queries to stimulate MLLMs' utilization of disciplinary and prior knowledge. As an MLLM benchmark, Creation-MMBench includes a rich variety of images to thoroughly evaluate multiple capabilities of MLLMs.}
    \label{fig:basic_statistics}
\end{figure*}

\noindent\textbf{Data Composition. } 
For each task, 15 carefully crafted test cases are collected. 
Each test case comprises two major components:

\begin{itemize}
\item \textbf{Visual Content}: One or more images that contain the necessary information required to accomplish the test case.
\item \textbf{Query}: Include Role (the identity models need to play), Background (prior knowledge that is not duplicated by the visual content and is difficult to acquire, Instruction (operations that models need to perform), and Requirement (constraints or additional considerations).
\end{itemize}

All queries are organized into a complete format using a unified template and sent to MLLMs with visual content. Instance-specified criteria are defined to make the evaluation more reasonable. 
The criteria can be mainly divided into two groups:

\begin{itemize}
\item \textbf{General Subjective Criteria}: Assess models' expressive capability (structure, style, fluency), execution ability for queries (compliance with requirements, roles, and instructions), and deep reflection on visual content.
\item \textbf{Visual Factuality Criteria}: Assess models' ability to perceive objective visual content and utilize visual information effectively.
\end{itemize}

\noindent\textbf{Data Annotation and Quality Control. } 
After task design and definition of data composition, we proceeded with data annotation (including questions and criteria) and quality control. To make the annotator easier to understand, we first built an example question for each task with detailed annotation, then asked volunteers to annotate 15 sample questions for each task with the example and guideline provided below:

\begin{enumerate}[leftmargin=*,topsep=0.5ex,itemsep=-0.5ex,partopsep=0.75ex,parsep=0.75ex,partopsep=0pt,wide,labelindent=0pt]

\item The \textbf{visual content} of questions should be semantic rich, and the query should not contain any explicit information in the visual content.

\item You are encouraged to formulate \textbf{diverse queries within the task scope}, like diverse roles and background settings, matching the visual content.

\item The ideal answer should be \textbf{open-ended, creative}, but the quality of the response can be assessed using criteria.

\item Ensure each \textbf{requirement is clear and avoids redundancy}. Keep the Visual Factuality Criteria concise and direct.

After initial labeling, we conducted cross-verification among volunteers, followed by expert review to ensure data quality. 

\end{enumerate}
\noindent\textbf{Evaluation Strategy.} We employ the MLLM-as-a-judge approach, which consists of two forms: Unitary Scoring and Pairwise Comparison.  In Unitary Scoring, the judging model assigns a score between 1 and 10 to the response of the evaluated model based on the Visual Factuality Criteria. The \textbf{Visual Factuality Score} is the average score across all questions. In Pairwise Comparison, the evaluated model is designated as model A, while the baseline model (GPT-4o-1120) is designated as model B. The judging model assesses the responses based on General Subjective Criteria and visual content, selecting from the set \{A\verb|>>|B, A\verb|>|B, A=B, A\verb|<|B, A\verb|<<|B\}.
To facilitate further computation, we assign numerical values to the pairwise comparison results: \{A\verb|>>|B = +2, A\verb|>|B = +1, A=B = 0, A\verb|<|B = -1, A\verb|<<|B = -2\}. For better interpretability, we multiply this average score by 50 and normalize it to the range of -100 to +100, forming a metric as \textbf{Reward}.  
To mitigate the inherent position bias in the MLLM-as-a-judge approach, we conduct a Dual Evaluation, swapping the response positions. The final result is obtained by averaging the outcomes of both evaluations. Detailed evaluation prompt is shown in Appendix B.

\subsection{Dataset Statistics}

\input{tables/main_table}

To better understand the composition of Creation-MMBench, we conducted a statistical analysis.  

\noindent\textbf{Benchmark Comparison}  ~\cref{tab: comparsion_with_partial_creation} shows the comparison result of Creation-MMBench and four widely used partial-creation MLLM benchmarks. As a dedicated benchmark for evaluating creativity, Creation-MMBench features a significantly richer set of creative questions and adopts a multi-image format. Each question is designed with specific roles to stimulate MLLMs' creative capabilities. Unlike other benchmarks that apply the same evaluation criteria across an entire benchmark or task, Creation-MMBench customizes assessment criteria for each question, taking into account both subjective creativity and visual factuality. This tailored approach enables a more comprehensive evaluation of MLLMs' creative abilities.

\noindent\textbf{Statistics and Cases}  \cref{fig:basic_statistics} presents several statistics and cases of Creation-MMBench. As depicted in \cref{fig:prompt_len}, we analyzed the query length distributions of Creation-MMBench in comparison with two partial-creation benchmarks (MLLM-Bench, AlignMMBench) and two widely used general benchmarks (MM-Vet, MMBench). The results indicate that our benchmark features more comprehensive and complex query designs. The majority of queries exceed a length of 500 tokens, which facilitates models in capturing richer creative contexts. \cref{fig:role_cloud} illustrates the diversity of roles present in the queries (e.g., writer, artist, Michelin chef, etc.), reflecting the richness of the questions. As an MLLM benchmark, our dataset contains a total of 1,001 images spanning more than 25 different categories, with some questions incorporating up to 9 images. \cref{fig:example_case} displays the example cases in Creation-MMBench.

\noindent\textbf{Vision Indispensability}  To verify the necessity of visual content in Creation-MMBench, we selected three MLLMs with varying capability levels (Gemini-1.5-Pro-002, Qwen2-VL-72B-instruct, and MiniCPM-o-2.6) and examined their performance after removing visual input. In \cref{fig:vision_indispensability}, we observe that when the visual information is removed, the same models exhibit significant declines in Reward. This finding verifies the necessity of visual content in evaluating model performance.

%% file: tables/main_table.tex
\newcolumntype{a}{>{\columncolor{gray!10}}c}
\newcolumntype{b}{>{\columncolor{gray!20}}c}
\newcolumntype{d}{>{\columncolor{gray!30}}c}
\newcolumntype{e}{>{\columncolor{gray!40}}c}
\begin{table*}[]
\centering
\small 
\resizebox{\textwidth}{!}{
\tablestyle{6pt}{1.4}
\begin{tabular}{l|cc|cc|cc|cc|cc|c|c}
\shline
\multirow{2}{*}{\bf Model} &  \multicolumn{2}{c|}{\bf Overall} &  \multicolumn{2}{c|}{\bf LW} &  \multicolumn{2}{c|}{\bf CFW} &  \multicolumn{2}{c|}{\bf PFW} & \multicolumn{2}{c|}{\bf CMU} & \multirow{2}{*}{\bf OC Score} & \multirow{2}{*}{\bf Avg Tokens}\\  \cline{2-11} 
 &  \textbf{VFS} & \textbf {Reward}  & \textbf{VFS} & \textbf {Reward} & \textbf{VFS} & \textbf {Reward} & \textbf{VFS} & \textbf {Reward} & \textbf{VFS} & \textbf {Reward} & &\\ \shline

\rowcolor{lightgray}
\multicolumn{13}{c}{\emph{Proprietary MLLMs }}  \\ \shline

Gemini-2.0-pro-exp & 8.53 & \bf 4.48 & \bf 8.66 & \bf -1.88 & 8.98& \bf 12.71 & 8.01 & \bf 3.33 & 8.65 & -8.06 & \bf 73.4 & \bf 718\\ 
\hdashline
\bf{GPT-4o-1120[Baseline]} & 8.72 & 0.00 & 8.86 & 0.00 & 8.93& 0.00 & 8.26 & 0.00 & 9.38 & 0.00 & 72.0 & 497\\ 
\hdashline
Gemini-1.5-pro-002 & 8.41 & -5.49 & \bf 8.66 & -6.04 & 8.59& -2.04 & \bf 8.05 & -4.82 & 8.75 & -17.22 & 72.2 & 444\\ 
GPT-4.5-0227 & \bf 8.54 & -5.88 &  8.63 & -4.38 & 8.76 & -8.33 & \bf 8.05 & -5.88 & \bf 9.29 & \bf -0.56 & / & 394\\
GPT-4o-mini & 8.07 & -13.56 &  8.30 & -4.38 & 8.44& -15.28 & 7.50 & -16.05 & 8.40 & -12.78 & 64.1 & 436\\ 
Doubao-VL & 8.38 & -14.09 &  8.28 & -19.17 & \bf 9.01& -3.33 & 7.65 & -18.72 & 8.77 & -25.00 & / & 516\\ 
Claude-3.5-Sonnet & 7.96 & -15.46 &  8.44 & -16.46 & 7.45& -21.57 & 7.98 & -11.14 & 8.88 & -9.44 & 70.6 & 336\\
Moonshot-v1-32k-vision & 7.43 & -20.58 &  7.30 & -21.46 & 8.20& -8.80 & 6.91 & -26.50 & 6.91 & -36.11 & / & 485\\\shline

\rowcolor{lightgray}
\multicolumn{13}{c}{\emph{Open-Source MLLMs}}  \\ \shline
Qwen2.5-VL-72B-Instruct & \bf 8.33 & \bf -5.82 &  8.04 & -10.83 & \bf 8.91& \bf 4.44 & \bf 7.68 & \bf -11.49 & \bf 8.86 & \bf -11.94 & 76.1 & \bf 553\\ 
InternVL2.5-78B-MPO & 8.06 & -12.55 &  \bf 8.22 & \bf -9.17 & 8.60& -5.00 & 7.45 & -16.32 & 8.22 & -27.78 & \bf 77.0 & 461\\ 
InternVL2.5-8B-MPO	&7.65	&-15.10	& 8.09	&-16.25&	8.30&	-3.80&	6.80 &	-23.95&	7.88	&-19.44&	70.3 & 548\\
InternVL2.5-78B	&7.91	&-16.43&	8.05	&-17.50	&8.45	&-7.69	&7.26	&-20.53&	8.18&	-28.33&	75.2 & 473\\
Qwen2-VL-72B-instruct&	7.87&	-22.45&	7.75	&-24.58	&8.17	&-15.56	&7.42	&-26.84	&8.43&	-26.39&	74.8 & 439\\
InternVL2.5-8B&	7.38&	-25.42&		7.91&	-23.33&	7.95&	-15.83&	6.62	&-33.95	&7.45&	-30.00&	68.1 & 500\\
Qwen2.5-VL-7B-Instruct&	7.55&	-29.80 &	7.34	&-39.38&	8.40 &	-21.67	&6.71	&-33.25&	7.78&	-30.56	&70.9 & 510\\
MiniCPM-o-2.6	&7.49&	-34.77	&7.79&	-35.42	&7.95&	-27.31&	6.76&	-40.88	&8.08&	-36.94	&70.2 & 389\\
DeepSeek-VL2&	7.24&	-38.52&	7.58&	-33.75&	7.58&	-32.50&	6.61	&-44.02	&7.81	&-45.56	&66.4 & 440\\
LLaVA-OneVision-72B	&7.16	&-39.87&		7.26&	-36.32	&7.72&	-30.61	&6.43	&-47.98&	7.62&	-46.37	&68.0 & 315\\
LLaVA-OneVision-7B	&6.75&	-43.49	&	7.36&	-43.54&	7.27&	-31.85	&6.04	&-50.53	&6.82&	-56.11&	60.2 & 373\\
Qwen2-VL-7B-instruct	&7.12	&-43.76& 6.99&	-55.83&	7.67	&-36.30&	6.57	&-45.26&	7.25&	-45.28&	67.1 & 456\\\shline
\end{tabular}
}
\caption{\textbf{Evaluation Result of MLLMs on Creation-MMBench.} VFS stands for Visual Factuality Score. LW, CFW, PFW, and CMU stand for four categories in Creation-MMBench: Literary Writing, Common Functional Writing, Professional Functional Writing, and Creative Multimodal Understanding. OC Score represents the average score of the OpenVLM Leaderboard and mainly demonstrates the objective performance of the model. The token number is calculated with tiktoken GPT-4o-1120 tokenizer.}
\label{tab:main-results}
\vspace{-4mm}
\end{table*}

%% file: sec/4_experiment.tex
\section{Experiment}
\label{experiment}

Using Creation-MMBench, we evaluate various Multimodal Large Language Models (MLLMs), with a focus on image-based MLLMs that support multiple image inputs. 
Additionally, we adapted our benchmark into a text-only version (Creation-MMBench-TO) by replacing the visual inputs with corresponding textual descriptions and tested multiple Large Language Models (LLMs) to gain deeper insights into their creative capabilities. 
All evaluations were conducted based on VLMEvalKit~\cite{duan2024vlmevalkit},
employing greedy decoding during inference with the maximum output tokens set to 4096.

\input{tables/language_model_table}

\subsection{Main Results}

We evaluated 20 current powerful MLLMs on Creation-MMBench, results are shown on ~\cref{tab:main-results}. 

\noindent\textbf{Proprietary MLLMs. } 
Gemini-2.0-Pro performs similarly to GPT-4o, particularly in common functional writing, where it excels in producing content with a conversational tone and effectively integrates images. Its strong pre-existing knowledge also helps in professional functional writing tasks, but there is a slight gap in perception, especially in tasks like document and snapshot analysis. The smaller GPT-4o-mini outperforms proprietary models like Claude but struggles with professional functional writing due to its limited disciplinary knowledge. DoubaoVL stands out in common functional writing tasks, achieving the highest visual factuality score in this area.

\noindent\textbf{Open-Source MLLMs. } 
Among open-source MLLMs, Qwen2.5-VL-72B stands out, performing similarly to advanced proprietary models like Gemini-1.5-Pro and outperforming GPT-4o-mini across all four major categories. This highlights the potential of open-source models in visual creation. The InternVL series also shows strong performance across different model sizes, indicating potential advantages in data and training strategies. The mixed preference optimized (MPO) model demonstrates impressive results in smaller models, with particular strengths in creative multimodal understanding, suggesting that MPO can effectively guide models to better align with human preferences.

\noindent\textbf{Category-level Evaluation Results. } Across all four categories, professional functional writing shows relatively weaker performance, while common functional writing performs the best. This may be due to the greater difficulty of tasks in the former, which require extensive disciplinary knowledge and a deeper understanding of image content. These tasks are more complex and demand higher cognitive abilities. In contrast, common functional writing typically involves simpler, everyday tasks that require less advanced image understanding, making them easier to complete. In the Multimodal Content Understanding and Creation category, while all models show basic content understanding, their ability to generate more creative content is limited. This highlights the gap between the models’ objective interpretation abilities and their human-aligned visual creativity, further qualitative cases are provided in Appendix G.

\begin{figure}[t]
\centering
\includegraphics[width=\linewidth]{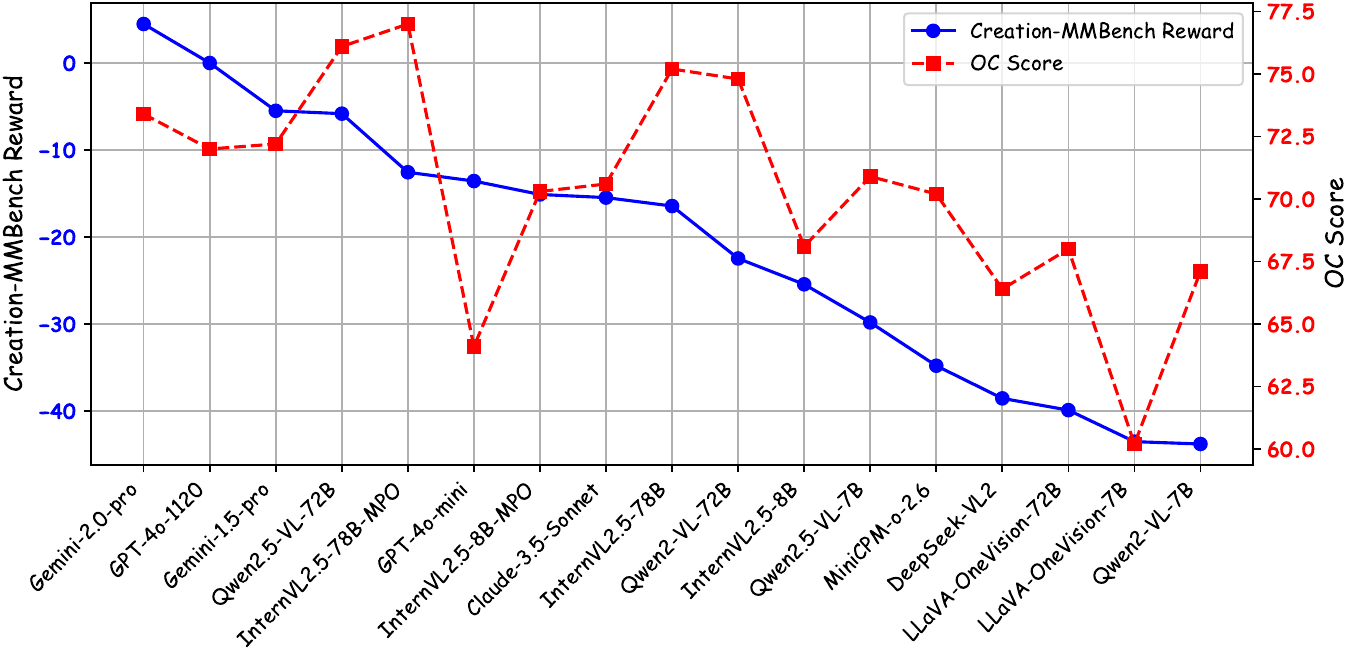}
\vspace{-2mm}
\caption{\textbf{Comparing OC Score and Creation-MMBench Reward.} This figure shows the model performance on the OpenVLM Leaderboard and Creation-MMBench, highlighting a significant gap between objective performance and visual creativity in some open-source models.}
\vspace{-4mm}
\label{fig:oc_score_diagram}
\end{figure}

\noindent\textbf{Comparison of Model Performance on Objective Tasks and Creation-MMBench.} To better compare the models’ objective performance with their visual creativity, we use the OC Score to represent the overall objective performance. As shown in ~\cref{fig:oc_score_diagram}, proprietary models perform well both in objective tasks and visual creativity. However, some open-source models, despite showing strong objective performance, struggle with open-ended visual creativity tasks. These models tend to excel in tasks with definitive answers but fall short in generating creative, contextually relevant content. This discrepancy emphasizes the need for a more comprehensive evaluation approach, as traditional objective metrics alone may not fully capture a model’s creative abilities in complex, real-world scenarios.

\subsection{Evaluating LLMs on Creation-MMBench-TO}

Current creation benchmarks for Large Language Models mostly focus on specific topics (e.g., LiveIdeaBench~\cite{ruan2024liveideabench}), but fail to reveal their creation capability in multiple daily scenarios. To investigate it, we build \textbf{Creation-MMBench-TO} and GPT-4o was used to make the image descriptions with the prompt shown in Appendix E. 
As shown in \cref{tab: VLM-LLM Result}, proprietary LLMs showed slightly better contextual creativity than open-source LLMs, though the gap was smaller than that between MLLMs. Large-scale language models performed better at understanding context and expressing ideas compared to smaller models. Additionally, the visual factuality score improved because GPT-4o’s image descriptions helped LLMs better interpret the image in comparison to MLLMs. Surprisingly, GPT-4o performed better in visual creativity on Creation-MMBench-TO. This could be because the model can focus more on divergent thinking and creation with the help of descriptions, which may minimize the negative impact of the basic visual content on creativity.
\subsection{Impact of Visual instruction tuning on creation capability of MLLM}

Existing research indicates that visual instruction tuning procedures may adversely affect the language encoder's capacity to process and model text-only inputs. 
To further investigate this, we conducted three experiments under different settings, as shown in \cref{tab: VLM-LLM Result}. The results indicate that the open-source MLLM, after visual instruction tuning, consistently performs worse compared to the corresponding LLM on Creation-MMBench-TO. 
This could be due to the instructions used during tuning being of similar length, which restricts the model’s ability to grasp detailed content in longer texts, resulting in a lower visual factuality score.
The lack of creative data that combines images further contributes to a significant drop in the reward score. 
Although some proprietary models have shown stronger performance on Creation-MMBench, the performance gap of most MLLMs on Creation-MMBench-TO and Creation-MMBench highlights the need for improvement in the perceptual capabilities of MLLMs.

\subsection{Evaluation Strategy Selection}
\input{tables/human_judge}

The goal of MLLM-as-a-judge is always to achieve a higher alignment with human preferences. Therefore, we randomly sampled a subset of questions (51 tasks × 2 questions) and recruited four volunteers to do the pairwise comparison. We selected three models (Gemini-1.5-pro-002, Qwen2-VL-72B, MiniCPM-o-2.6) as Model A, used the baseline model (GPT-4o-1120) as Model B, randomizing the responses' position to avoid human biases. Details of the human evaluation process are provided in Appendix F.

We then selected three advanced MLLMs (Gemini-2.0-Pro, Claude-3.5-Sonnet, GPT-4o) as judging models, and used MAE and Consistency as metrics to reflect the alignment degree. \cref{tab: model-human-alignment} presents the alignment degree between different evaluation strategies and human preferences. The results indicate that for all judging models, Dual Evaluation outperforms Single Evaluation, verifying the necessity of Dual Evaluation. Among all the judging models, GPT-4o achieves the best performance in terms of MAE and Consistency, exhibiting the highest alignment with human preferences. 
Finally, we selected Dual Evaluation, and GPT-4o as the evaluation strategy for Creation-MMBench.

\subsection{Qualitative Study}

\begin{figure}[t]
\centering
\includegraphics[width=\linewidth]{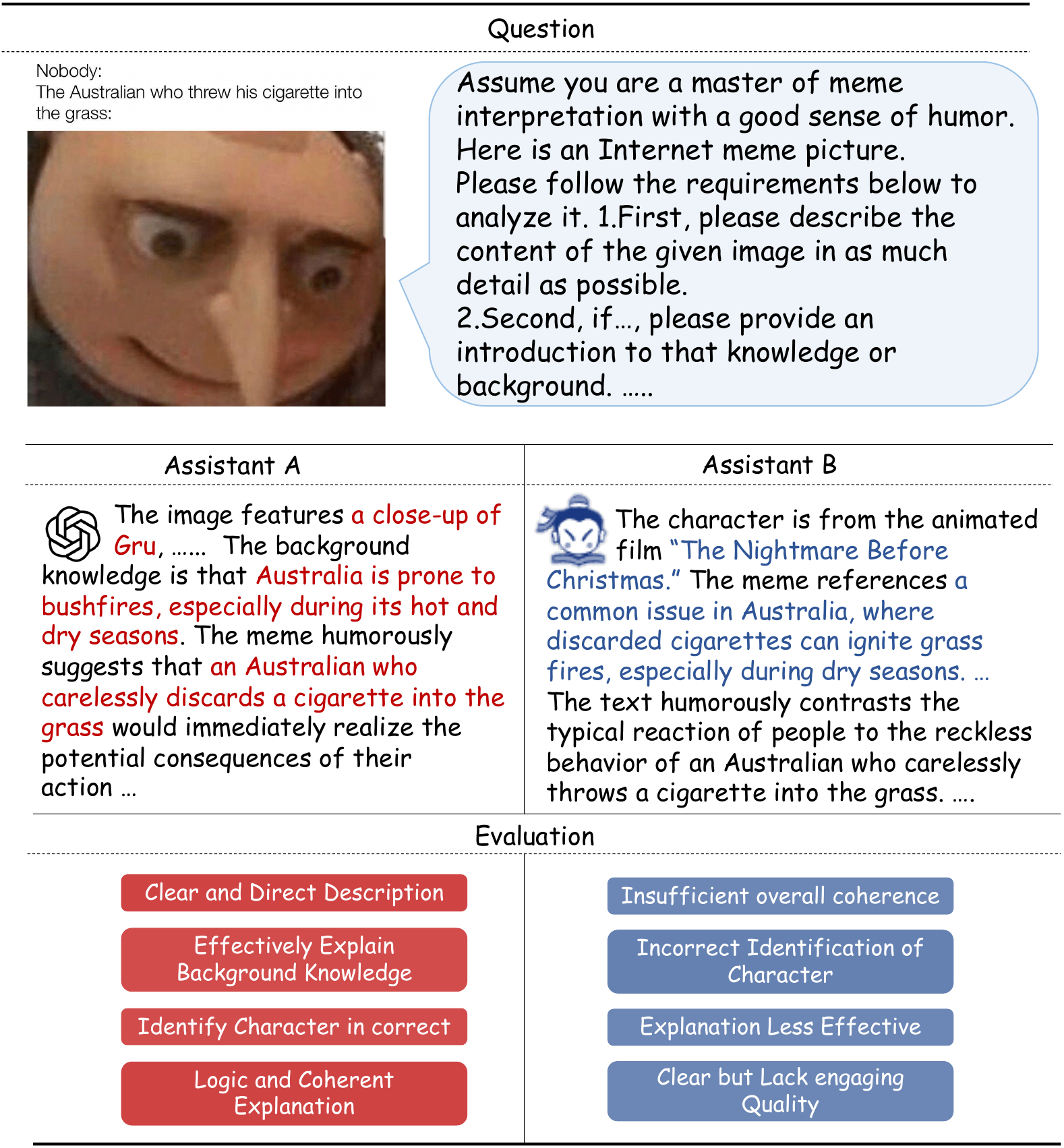}
\vspace{-2mm}
\caption{\textbf{Qualitative study Case between InternVL-2.5-78B and Reference Answer (GPT4o-1120).} }
\vspace{-4mm}
\label{fig:qualitative_study}
\end{figure}

To further explore the differences between models on Creation-MMBench, we conducted a detailed qualitative study by combining model responses with evaluations. As shown in ~\cref{fig:qualitative_study}, InternVL2.5 exhibited limitations in visual perception, particularly in accurately identifying characters due to insufficient latent knowledge. Additionally, InternVL2.5 showed certain weaknesses in the fluency and engagement of its language expression. In contrast, GPT-4o was favored by the evaluation model, which provided a more balanced assessment. This highlights that open-source models still have considerable space for improvement, particularly in visual creativity tasks.

%% file: tables/language_model_table.tex
\begin{table*}[t]
\centering
\tablestyle{8pt}{1.1}
\resizebox{\textwidth}{!}{%
\begin{tabular}{l|l|cc|cc|cc}
\shline
\multirow{2}{*}{\bf VLM} & \multirow{2}{*}{\bf Corresponding LLM} & \multicolumn{2}{c|}{\bf Text Input w. LLM} & \multicolumn{2}{c|}{\bf Text Input w. VLM} & \multicolumn{2}{c}{\bf Vision+Text Input w. VLM}\\ \scline{3-8}
& & \textbf {VFS} & \textbf {Reward}  & \textbf {VFS} & \textbf {Reward}  & \textbf {VFS} & \textbf {Reward}  \\ \shline
GPT-4o-1120	& GPT-4o-1120 & \bf 8.71 & \bf 6.96 &  \bf 8.71 & \bf 6.96  & \bf 8.72 & 0.36  \\ 
Gemini-2.0-pro-exp	& Gemini-2.0-pro-exp & 8.49 & 4.08  & 8.49 & 4.08  & 8.53 & \bf 4.48   \\ 
Qwen2.5-VL-72B-Instruct	& Qwen2.5-72B-Instruct & 8.55 & 0.82 & 8.51 & -4.05 & 8.33 & -5.82  \\ 
Qwen2.5-VL-7B-Instruct	& Qwen2.5-7B-Instruct & 8.18 & -19.18 & 7.97 & -27.50 & 7.55 & -29.80  \\ 
MiniCPM-o-2.6 & Qwen2.5-7B-Instruct &	8.18 & -19.18  & 7.78 & -36.57  & 7.49 & -34.77 \\
InternVL2.5-8B &	InternLM2.5-7B-Chat &	7.83 & -22.19 & 7.92 & -28.73  & 7.38 & -25.42\\ \shline

\end{tabular}
}

\caption{\textbf{LLM performance on Creation-MMBench-TO and Visual Instruction Tuning Impact on VLM creation capability.} The image descriptions provided by GPT-4o are general. For the proprietary models, we point to themselves as corresponding LLM and report the performance with image descriptions and questions.}
\label{tab: VLM-LLM Result}
\end{table*}

%% file: tables/human_judge.tex
\begin{table}[t]
\centering
\tablestyle{4pt}{1.1}
\resizebox{\linewidth}{!}{%

\begin{tabular}{c|c|c|c|c|c|c|c|c|c}
\shline
\multirow{2}{*}{\bf Judger} & \multirow{2}{*}{\bf MLLM} & \multicolumn{4}{c|}{\bf Dual Eval} & \multicolumn{4}{c}{\bf Single Eval} \\ \scline{3-10}
& & \multicolumn{2}{c|}{\textbf{MAE$\downarrow$}} & \multicolumn{2}{c|}{\textbf{Cons.$\uparrow$}} & \multicolumn{2}{c|}{\textbf{MAE$\downarrow$}} & \multicolumn{2}{c}{\textbf{Cons.$\uparrow$}} \\ \shline
\multirow{3}{*}{Gemini-2P} & Gemini & 0.65 & \multirow{3}{*}{0.59} & 82.83 & \multirow{3}{*}{86.67} & 0.78 & \multirow{3}{*}{0.72} & 74.75 & \multirow{3}{*}{78.67} \\
& Qwen & 0.51 &  & 91.00 &  & 0.67 &  & 80.00 &  \\
& MiniCPM & 0.61 &  & 86.14 &  & 0.69 &  & 81.19 &  \\ \shline
\multirow{3}{*}{Claude-3.5} & Gemini & 0.56 & \multirow{3}{*}{0.50} & 89.90 & \multirow{3}{*}{90.60} & 0.61 & \multirow{3}{*}{0.59} & 83.84 & \multirow{3}{*}{85.23} \\
& Qwen & 0.46 &  & 92.00 &  & 0.59 &  & 85.00 &  \\
& MiniCPM & 0.47 &  & 89.90 &  & 0.57 &  & 86.87 &  \\ \shline
\multirow{3}{*}{\textbf{GPT-4o}} & Gemini & 0.53 & \multirow{3}{*}{\textbf{0.50}} & 92.08 & \multirow{3}{*}{\textbf{92.13}} & 0.57 & \multirow{3}{*}{\textbf{0.54}} & 89.11 & \multirow{3}{*}{\textbf{88.85}} \\
& Qwen & 0.42 &  & 96.08 &  & 0.46 &  & 91.18 &  \\
& MiniCPM & 0.53 &  & 88.24 &  & 0.59 &  & 86.27 &  \\ \shline
\end{tabular}
}
\caption{\textbf{The Alignment Between Different Evaluation Strategies and Human Preference.} }
\label{tab: model-human-alignment}
\end{table}

%% file: sec/6_conclusion.tex
\section{Conclusion}
\label{conclusion}

We present Creation-MMBench, a novel benchmark designed to assess the creative capabilities of MLLMs in real-world scenarios. The benchmark consists of 765 cases across 51 detailed tasks. For each case, we develop instance-specific criteria to evaluate both the subjective quality of responses and visual-factual alignment. Additionally, we create a text-only version, Creation-MMBench-TO, by substituting image inputs with corresponding textual descriptions. Extensive experiments on both benchmarks enable a thorough assessment of mainstream MLLMs’ creative abilities and allow us to examine the negative impact of visual instruction tuning.

%% file: sec/X_suppl.tex
\clearpage
\setcounter{page}{1}
\maketitlesupplementary

\begin{figure*}[t]
\centering
\includegraphics[width=\linewidth]{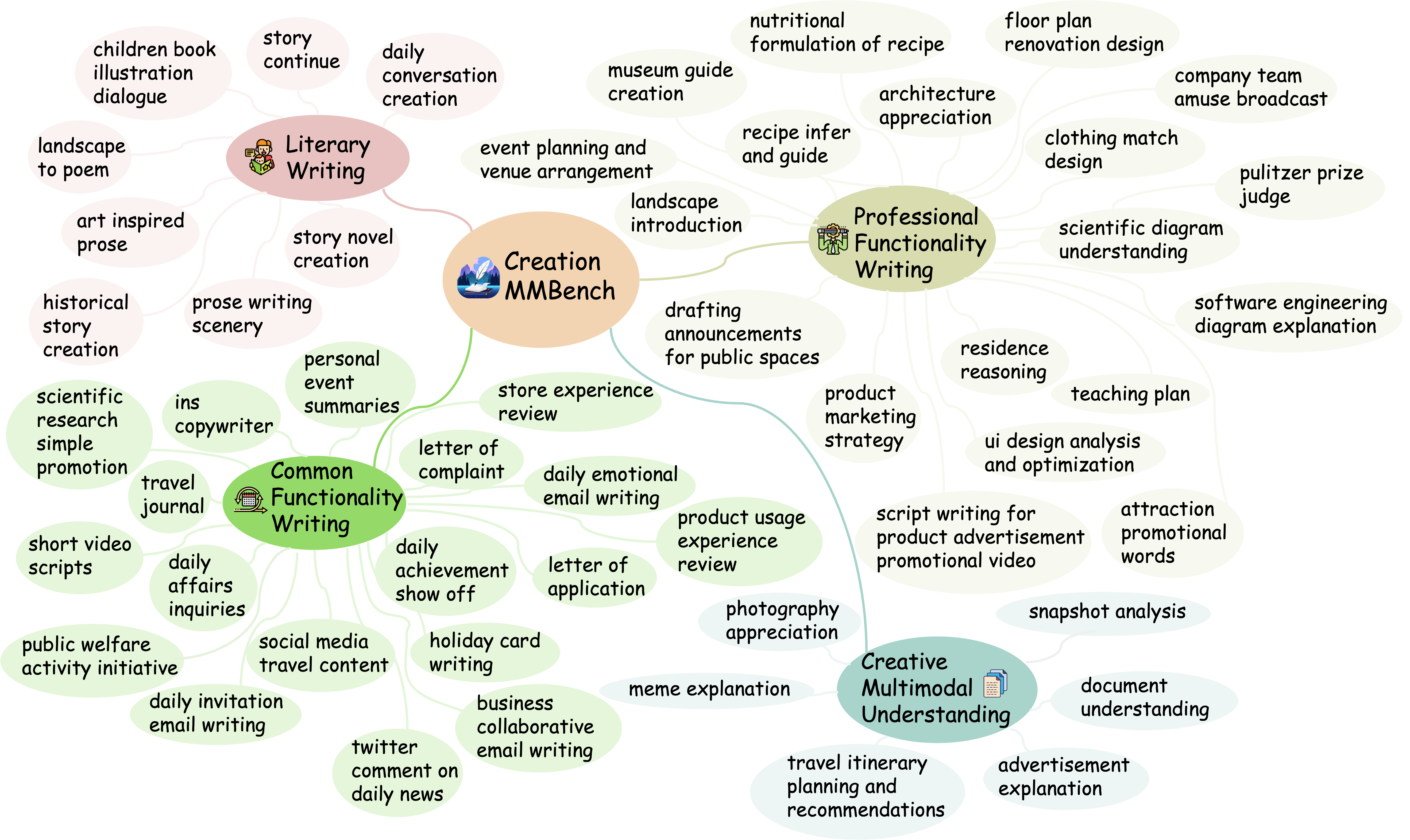}
\caption{\textbf{Overview of Creation-MMBench Complete Task.} Contains four task categories, each category consists of multiple tasks. }
\label{fig:task-full-overview}
\end{figure*}

\section*{A Overview of Tasks in Creation-MMBench}
\label{appendix_tasks}

Creation-MMBench consists of four main categories and 51 tasks, as shown in ~\cref{fig:task-full-overview}. The Literary Writing category includes 8 tasks, focusing on visual literary creation using images such as photographs, illustrations, and paintings. The Common Functional Writing category comprises 18 tasks, addressing everyday functional creation across various genres and image types. The Professional Functional Writing category contains 19 tasks, focusing on creation tasks that require specific domain knowledge. Finally, the Creative Multimodal Understanding category includes 6 tasks, which involve interpreting implied content from images with rich textual information. For each category, we provide four examples, as illustrated in \cref{fig:LW_story_continue} - \cref{fig:CMU_travel_itinerary_planning_and_recommendations}.

\section*{B Query and Judge Prompt Template for Creation-MMBench}
\label{appendix_prompt}

\subsection*{B.1 Query Template}

For each test case, the query is formatted using the template shown in ~\cref{fig:Question_prompt_cmb}. In Creation-MMBench-TO, we replace visual content with generated descriptions, as no images are provided to the LLM.

\subsection*{B.2 Judge Template}

For pairwise comparison, the General Subjective Criteria are essential for a fair assessment. Visual content helps the judge model better understand the predictions and prevents arbitrary conclusions based solely on linguistic strengths. As shown in ~\cref{fig: sub_judge_creation}, the predictions from different models are presented side by side with the criteria to minimize position bias, with instructions also provided to the judging model. Although changing the hypothetical positions helps reduce positional bias, dual evaluation remains necessary. The format restrictions for evaluating model responses facilitate the extraction of the final verdict through regular matching methods.

For Unitary Scoring, we provide Visual Factuality Criteria along with the model’s predictions, the reference answer, and the query, as outlined in ~\cref{fig: vf_judge_creation}. In test cases with a corresponding GroundTruth, this is included to ensure accurate judgment. Each criterion includes several main points, which may be further subdivided into subpoints. The evaluation model scores based on the completeness of these points, with a total score of 10.

\section*{C Main Experiment Analysis on Win Rate}
\label{appendix_main_exp}

In Creation-MMBench, we adopt the MLLM-as-a-judge approach and introduce two metrics, Visual Factuality Score and Reward, to assess the creative capabilities of MLLMs. In this section, we propose a new metric, \textbf{Win Rate}, to provide a more comprehensive evaluation of MLLMs' performance.

\input{tables/win_rate}

\subsection*{C.1 Win Rate Definition}
\label{appendix_c_1}

\textbf{Win Rate} is defined as the proportion of instances in which the response generated by the evaluated model surpasses that of the baseline model in the Pairwise Comparison.

\subsection*{C.2 Main results on Win Rate}
\label{appendix_c_2}
~\cref{tab:win_rate_table} presents the Win Rate and detailed judgment counts of MLLMs on Creation-MMBench. 
Among Proprietary MLLMs, Gemini-2.0-pro-exp demonstrates the best performance in terms of Win Rate, exhibiting the highest number of Much Better and Better cases. In contrast, GPT-4o-mini performs the worst, with only 53 Better cases.  
Among Open-Source MLLMs, Qwen2.5-VL-72B-Instruct achieves the best performance, with only 42 Much Worse cases. However, most models perform poorly, lacking any Much Better cases.  
A noticeable performance gap remains between Open-Source and Proprietary MLLMs in terms of Win Rate.

\section*{D Advanced Analysis of Creation-MMBench}
\label{appendix_d}

\subsection*{D.1 Redundancy Analysis}
\label{appendix_d_1}

\begin{figure*}[htbp]
    \centering

    \begin{subfigure}{0.49\textwidth}  
        \includegraphics[width=\linewidth]{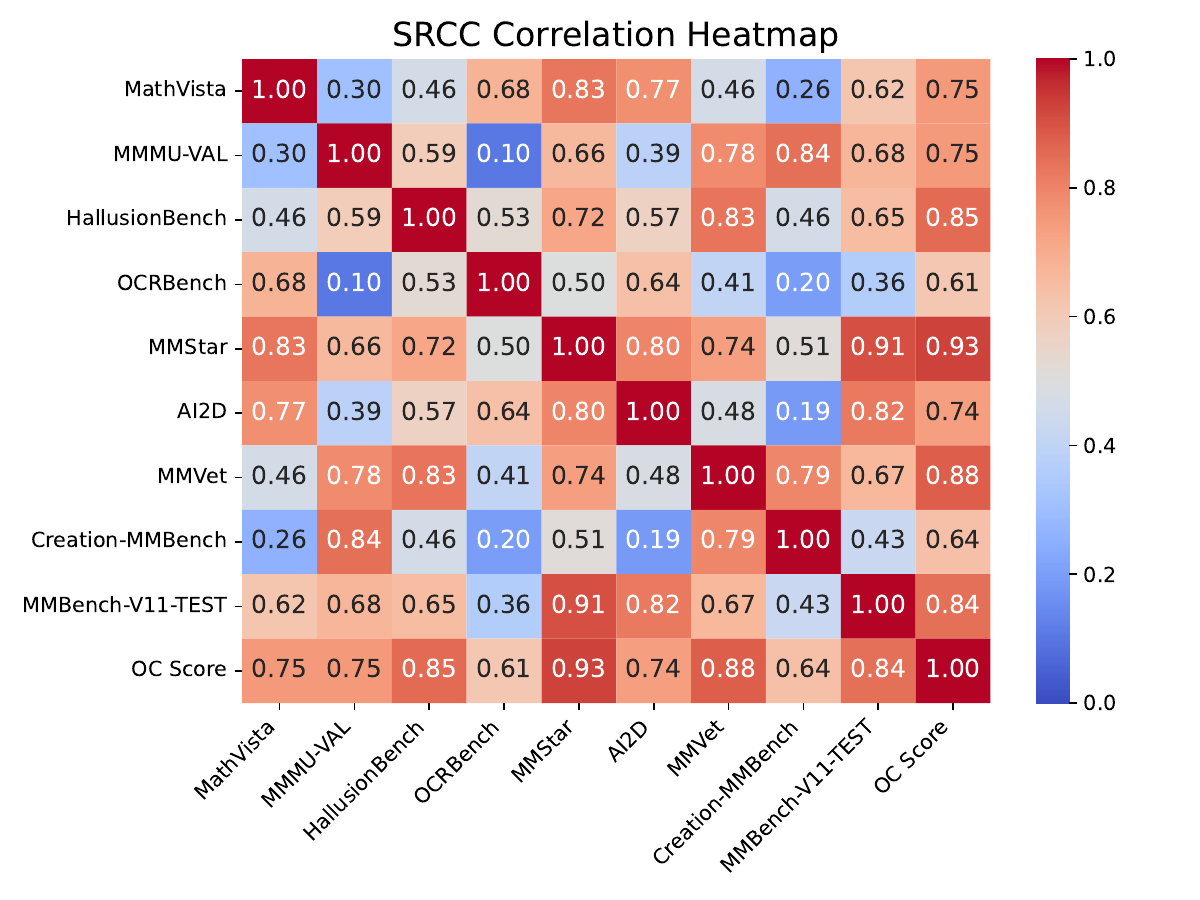}
        \caption{Distribution of query lengths.}
        \label{fig:srcc}
    \end{subfigure}
    \begin{subfigure}{0.49\textwidth}  
        \includegraphics[width=\linewidth]{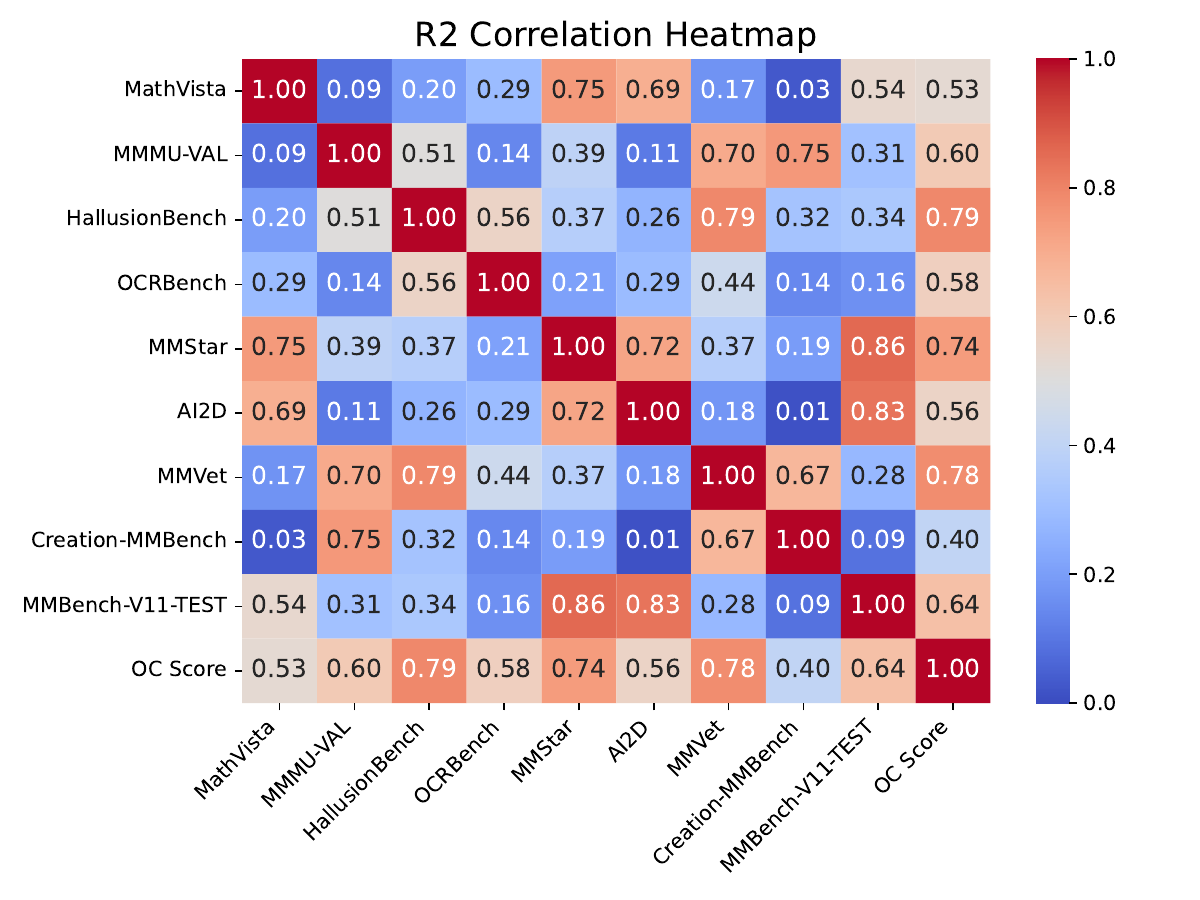}
        \caption{Roles in Creation-MMBench.}
        \label{fig:r2}
    \end{subfigure}

    \caption{\textbf{Redundancy Analysis of Creation-MMBench with other widely used MLLM Benchmarks.} }
    \label{fig:redundancy}
\end{figure*}

Following the ~\cite{zhang2025redundancyprinciplesmllmsbenchmarks}, we compute the correlation coefficients between the model evaluation results of Creation-MMBench and other representative objective benchmarks to investigate the redundancy of Creation-MMBench. ~\cref{fig:redundancy} presents the Spearman’s Rank Correlation Coefficient (SRCC) and the coefficient of determination (R²) between the benchmarks. As shown in the figure, Creation-MMBench exhibits a low correlation with MathVista, AI2D, and OCRBench in both SRCC and R². This is likely because these three benchmarks primarily assess objective capabilities such as mathematical reasoning, information extraction, and simple logical inference, with most queries presented in multiple-choice format—an evaluation focus that differs significantly from that of Creation-MMBench.

In contrast, MMMU and MM-Vet show a certain degree of correlation with Creation-MMBench. This may be attributed to the fact that both benchmarks incorporate a portion of creativity-oriented testing, such as the Art \& Design section in MMMU-Val and the summarization task in MM-Vet. In general, Creation-MMBench shows low redundancy with existing MLLM Benchmarks, which reflects the novelty and uniqueness of our benchmark.


\subsection*{D.2 Other Statistics}
\label{appendix_d_2}
\begin{figure*}[htbp]
    \centering
    \begin{subfigure}{0.5\textwidth}  
        \centering
        \includegraphics[width=\textwidth]{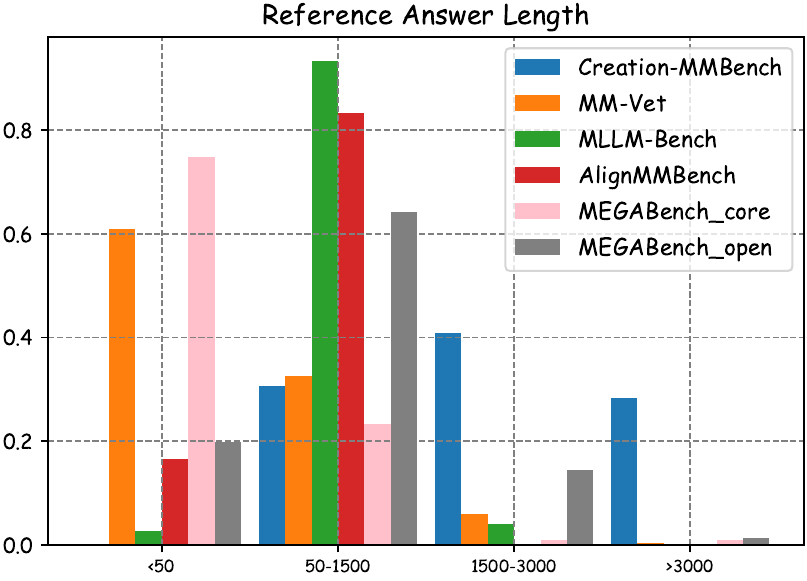}
        \caption{Distribution of reference answers lengths.}
        \label{fig:refer_len}
    \end{subfigure}
    \begin{subfigure}{0.5\textwidth}  
        \centering
        \includegraphics[width=\textwidth]{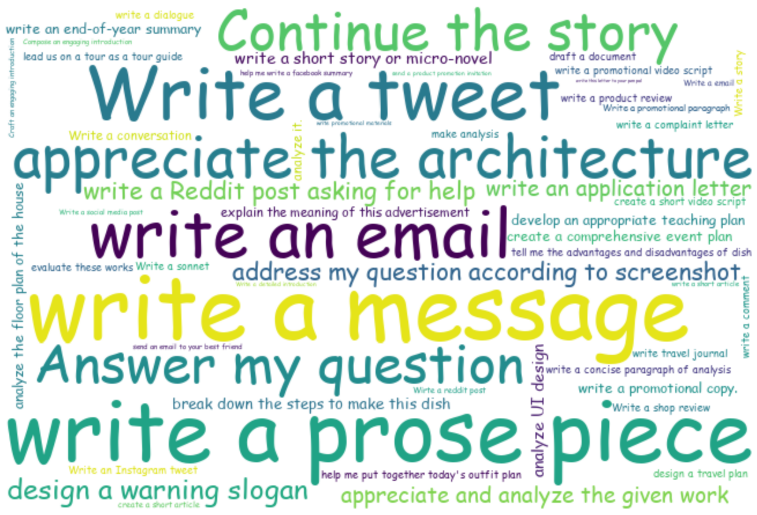}
        \caption{Instructions in Creation-MMBench.}
        \label{fig:inst_cloud}
    \end{subfigure}
    \begin{subfigure}{0.8\textwidth}  
        \centering
        \includegraphics[width=\textwidth]{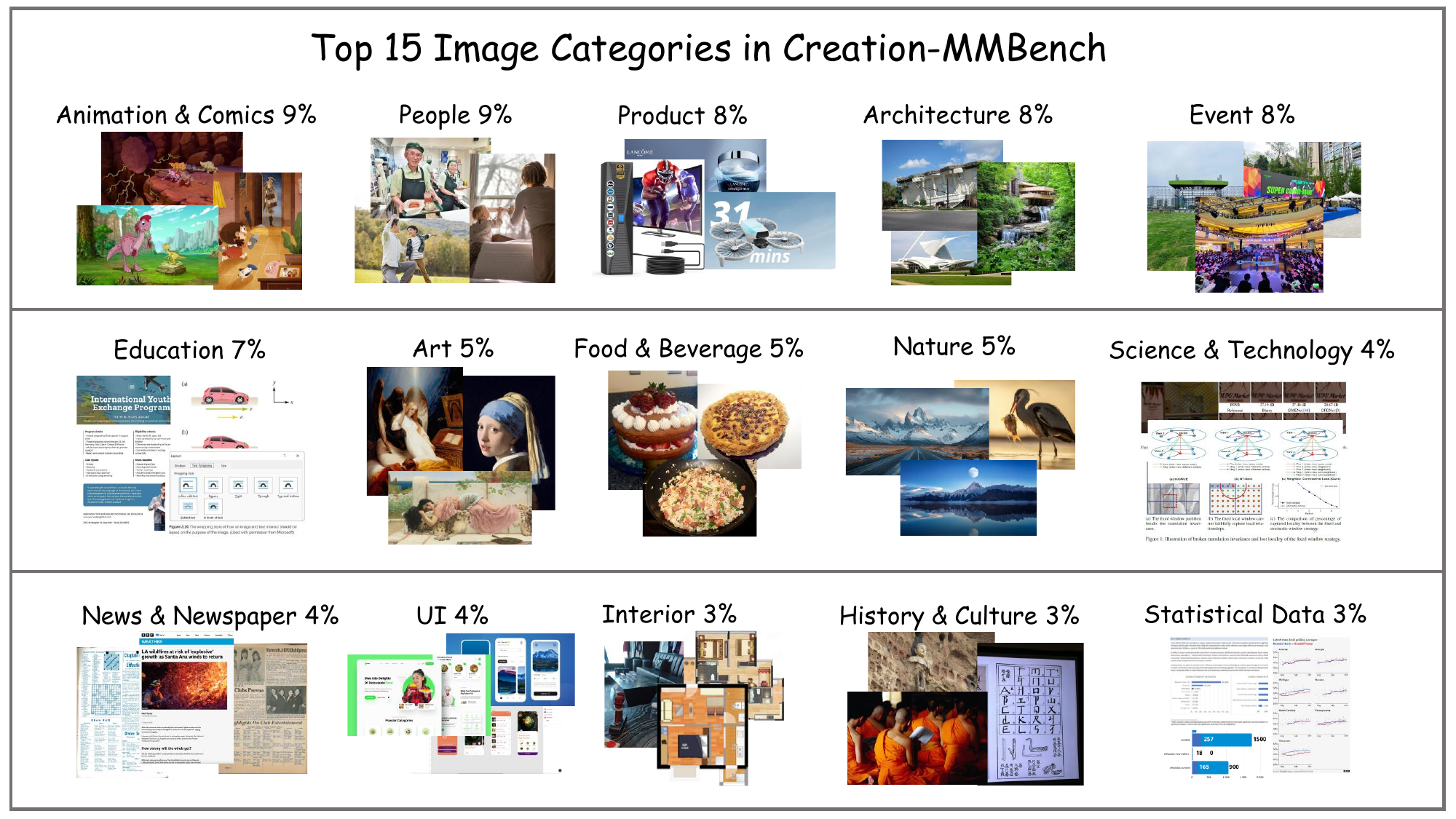}
        \caption{Top 15 Image Categories in Creation-MMBench.}
        \label{fig:top_15_images}
    \end{subfigure}
    \caption{\textbf{Other Statistics of Creation-MMBench.}}
    \label{fig:other_statistics}
\end{figure*}

~\cref{fig:other_statistics} presents supplementary statistics for Creation-MMBench. As shown in ~\cref{fig:refer_len}, we compare the reference answer lengths of Creation-MMBench with four widely used MLLM benchmarks. It is evident that our benchmark exhibits a significantly higher proportion of long answers exceeding 1,500 tokens, which reflects the greater complexity of our tasks. ~\cref{fig:inst_cloud} illustrates the richness of instructions within Creation-MMBench, reflecting the diversity of tasks. The analysis of image categories in ~\cref{fig:top_15_images} demonstrates the rich visual content incorporated in our benchmark. This diversity ensures a comprehensive evaluation of the model’s perceptual capabilities, further solidifying Creation-MMBench as a rigorous MLLM benchmark.

\section*{E Query-Specific Experiments on Creation-MMBench-TO}
\label{appendix_e}

For Creation-MMBench-TO, the instructions for visual content description are crucial as they are designed to fully stimulate the model to interpret the content of the image as detailed and rich as possible. To avoid the loss of some fine-grained content caused by generic visual descriptions, which could affect the performance of LLM's creative ability, we additionally used Query-Specific Instruction generated by GPT-4o to guide the visual description~\cite{qiao2025prism}.

As shown in ~\cref{fig:description_prompt}, Generic instruction is a standardized, universal instruction aimed at extracting and describing the basic elements present in an image. Query-specific instruction is a combination of generic instruction and incremental instruction that directs the VLM to provide a detailed account of the visual information relevant to the
question. The incremental instruction is crafted by the GPT-4o given the text-only question and the few-shot prompt template shown in ~\cref{fig:few_shot_prompt}.

\input{tables/description_comparison_table}
Results on ~\cref{tab: generic-query_specific comparison} reveal that query-specific descriptions can help LLMs gain a better understanding of visual content, resulting in a higher Visual Factuality Score and Reward. However, GPT-4o exhibits an inverse trend, which may be because fine-grained descriptions can mislead the attention of the models and may generate too much detailed creative content that does not fully meet the criteria.


\section*{F Human Alignment}
\label{appendix_f}

In this section, we provide a detailed examination of Human Alignment, covering the process of pairwise comparison conducted by human evaluators, the definition of the evaluation metrics, and the comprehensive results of Model-Human and Human-Human alignment.

\subsection*{F.1 The process of Human Pairwise Comparison}
\label{appendix_f_1}

\begin{figure*}[htbp]
    \centering
    \includegraphics[width=\textwidth]{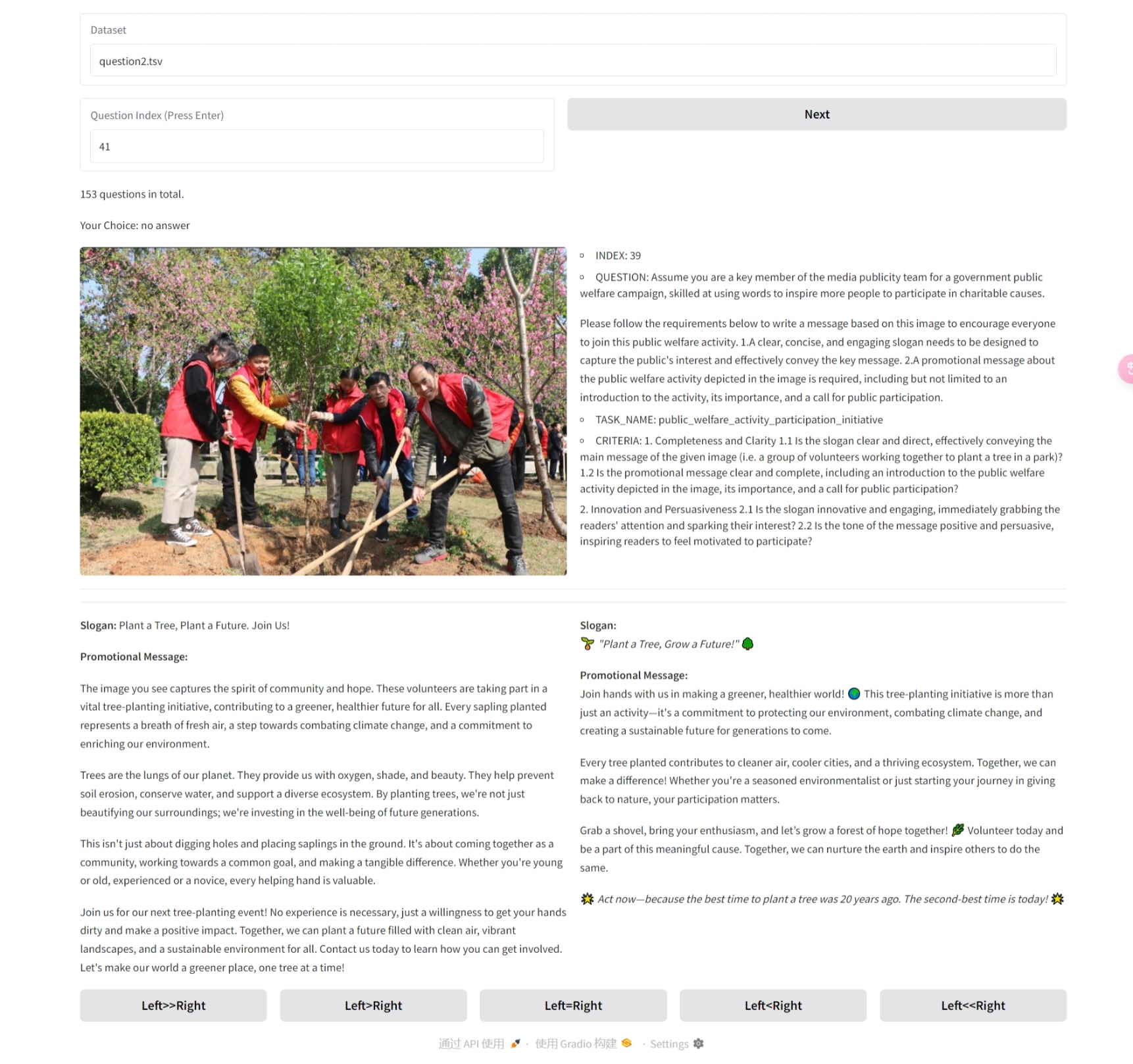}
    \caption{\textbf{The Process of Human Pairwise Comparison.}}
    \label{fig:data_voter}
\end{figure*}

For human evaluation, We sampled two questions from each task in Creation-MMBench to construct a set of 102 questions. Four volunteers were recruited to perform pairwise comparisons on this question set. ~\cref{fig:data_voter} illustrates the user interface used by human evaluators for this task. To mitigate potential bias, we randomized both the order of the questions and the positions of Model A (Gemini-1.5-pro-002, Qwen2-VL-72B, MiniCPM-o-2.6) and Model B (baseline, i.e. GPT-4o-1120)’s responses. Evaluators were provided with the corresponding visual content, related questions, and assessment criteria to compare the quality of the responses presented on the left and right. Their selections were recorded to generate preference results.

\subsection*{F.2 The Definition of MAE and Consistency}
\label{appendix_f_2}

Eq ~\eqref{eq:mae} and ~\eqref{eq:consistency} present the metrics used to evaluate the degree of alignment, specifically MAE and Consistency. In these equations, $\mathcal{J}$ represents the pairwise comparison results from a specific judging model or human evaluator, while $\mathcal{P}$ denotes the corresponding reference value (average of human ratings).

\begin{equation}
    \textbf{MAE} = \frac{1}{n} \sum_{i=1}^{n} | \mathcal{J}_i - \mathcal{P}_i |
    \label{eq:mae}
\end{equation}

\begin{equation}
    \textbf{Consistency} = \frac{1}{n} \sum_{i=1}^{n} 
    \begin{cases} 
        1, & \text{if } |\mathcal{J}_i - \mathcal{P}_i| \leq 1  \\
        0, & \text{otherwise}
    \end{cases}
    \label{eq:consistency}
\end{equation}

\subsection*{F.3 Full Results}
\label{appendix_f_3}

We conducted experiments to study both Model-Human Alignment and Human-Human Alignment. For the former, $\mathcal{J}$ refers to the judging model's comparison result, while $\mathcal{P}$ represents the average human preference. For the latter, $\mathcal{J}$ refers to the comparison result of an individual human, with $\mathcal{P}$ being the average preference of the remaining humans. ~\cref{tab:model-human-alignment} presents the detailed alignment results.  

It can be observed that for all judging models, MLLM-as-a-judge outperforms LLM-as-a-judge in terms of MAE and Consistency. This may be because the incorporation of visual content allows the judging models to conduct a more comprehensive evaluation.  Regarding Human-Human Alignment, human preferences are not highly consistent with one another, which reflects the subjective nature of our benchmark.

\input{tables/human_judge_2}

\section*{G Category Qualitative Case Study}

\begin{figure*}[t]
\centering
\includegraphics[width=\linewidth]{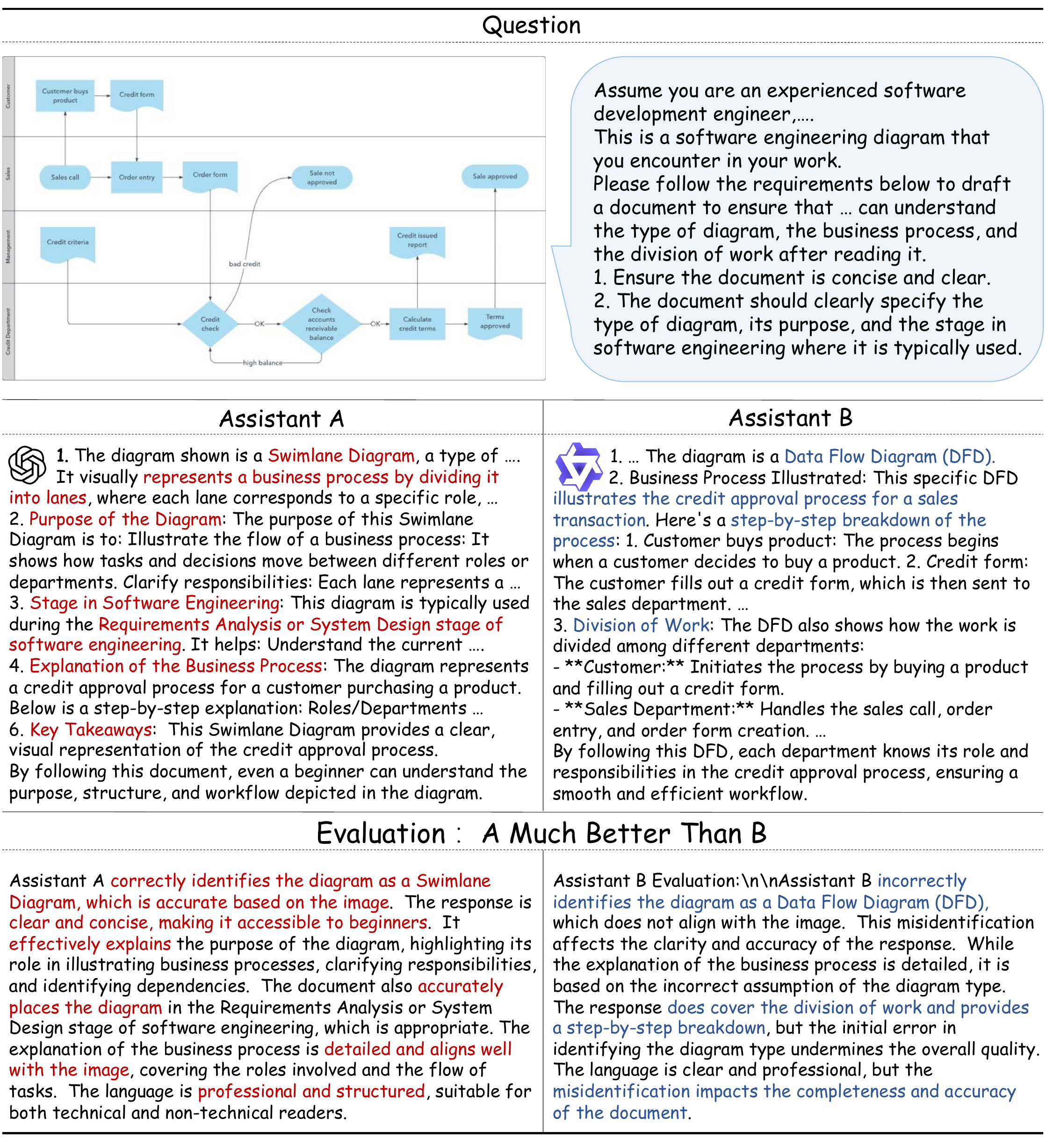}
\caption{\textbf{Qualitative Case in Professional Functional Writing.} This case comes from Software Engineering Diagram Explanation Task, Assistant A is GPT-4o-1120, assistant B is Qwen2.5-VL-72B.}
\label{fig:qualitative_case_appendix_1}
\end{figure*}

We conducted a qualitative analysis of the common situations that occur in some task categories.
~\cref{fig:qualitative_case_appendix_1} mainly focuses on the category of Professional Functional Writing. It can be significantly observed that Qwen2.5-VL misjudged the swimlane diagram as a data flow diagram due to insufficient understanding of the domain-specific knowledge, leading to subsequent errors in diagram analysis. In contrast, GPT-4o-1120 effectively avoided this mistake, and its overall language is more professional and structured, demonstrating a more accurate and detailed explanation of the diagram, thus gaining the preference of the judge model. This example also reflects the important role of specific disciplinary knowledge and a detailed understanding of image content in this category of tasks.

\begin{figure*}[t]
\centering
\includegraphics[width=\linewidth]{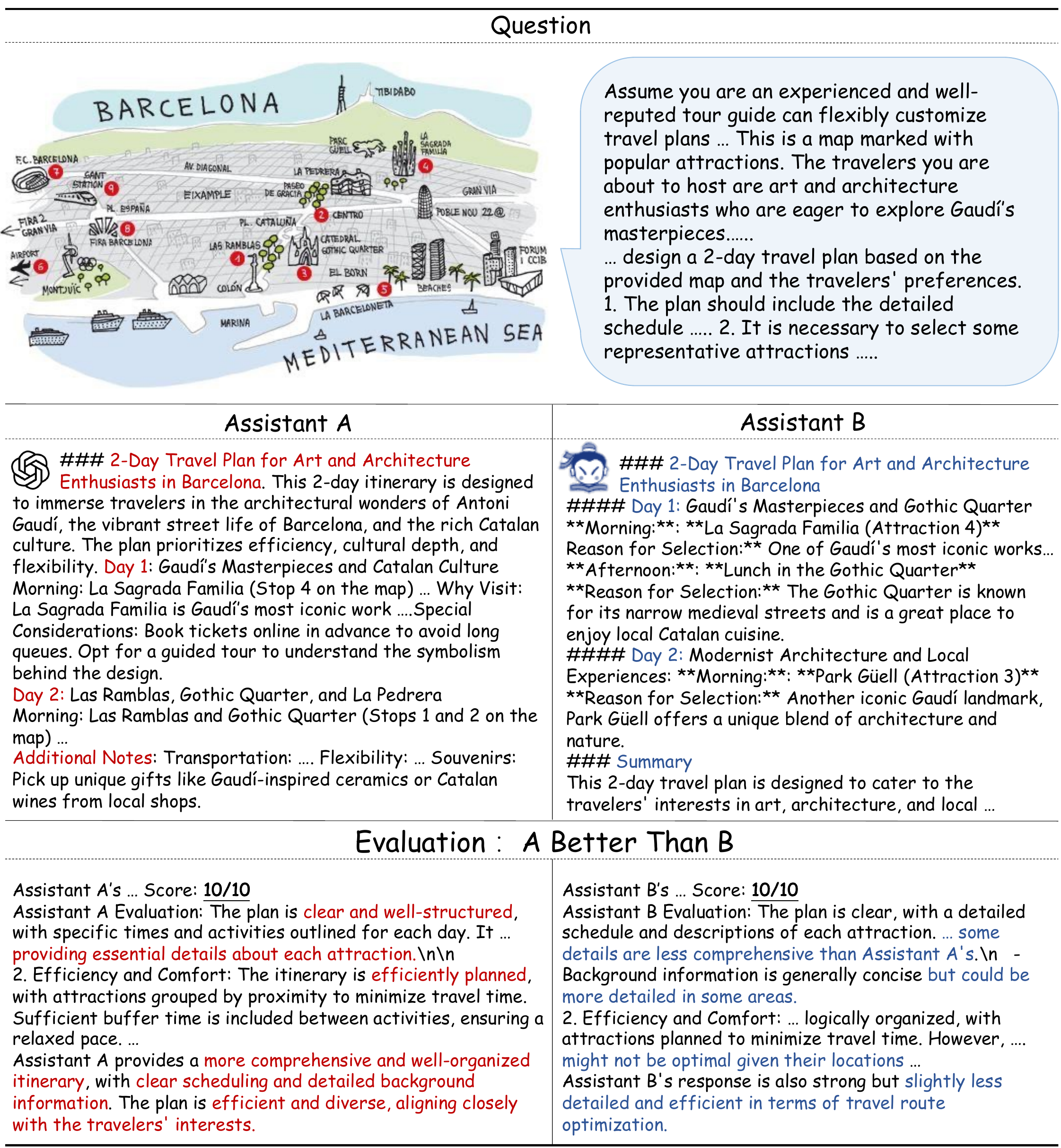}
\caption{\textbf{Qualitative Case in Creative Multimodal Understanding.} This case comes from Travel Itinerary Planning and Recommendations Task, Assistant A is GPT-4o-1120, assistant B is InternVL2.5-78B.}
\label{fig:qualitative_case_appendix_2}
\end{figure*}

For Creative Multimodal Understanding tasks, as shown in ~\cref{fig:qualitative_case_appendix_2}, both models gain full scores in visual factuality and exhibit similar performance in basic visual content understanding and information extraction. However, GPT-4o-1120 gives a more comprehensive plan with clear scheduling and reasonable arrangement, thus winning the preference of the judging model. 

\input{tables/category_case}

\input{tables/prompt_template}

%% file: tables/win_rate.tex
\newcolumntype{a}{>{\columncolor{gray!10}}c}
\newcolumntype{b}{>{\columncolor{gray!20}}c}
\newcolumntype{d}{>{\columncolor{gray!30}}c}
\newcolumntype{e}{>{\columncolor{gray!40}}c}
\begin{table*}[]
\centering
\small 
\resizebox{\textwidth}{!}{
\tablestyle{4pt}{1.2}
\begin{tabular}{l|ccccccccc}
\shline
\textbf{Model} & \textbf{VFS} & \textbf{Reward} & \textbf{WR} & \textbf{MB} & \textbf{Better} & \textbf{Tie} & \textbf{Worse} & \textbf{MW} & \textbf{Fail}\\ \shline

\rowcolor{lightgray}
\multicolumn{10}{c}{\emph{Proprietary MLLMs }}  \\ \shline

Gemini-2.0-pro-exp & 8.53 & 4.48 & 26.75$\%$ & 9 & 400 & 898 & 163 & 59 & 1 \\
\bf{GPT-4o-1120} & 8.72 & 0.00 & - & - & - & -  & - & - & - \\
Gemini-1.5-pro-002 & 8.41 & -5.49 & 11.37$\%$ & 6 & 168 & 1032 & 300 & 24 & 0 \\
GPT-4.5-0227 & 8.54 & -5.88 & 5.36$\%$ & 7 & 75 & 1186 & 255 & 7 & 0 \\
GPT-4o-mini & 8.07 & -13.56 & 3.79$\%$ & 5 & 53 & 1022 & 422 & 28 & 0 \\
Doubao-VL & 8.38 & -14.09 & 9.22$\%$ & 4 & 137 & 850 & 500 & 38 & 1 \\
Claude-3.5-Sonnet & 7.96 & -15.46 & 12.55$\%$ & 4 & 188 & 843 & 321 & 174 & 0 \\
Moonshot-v1-32k-vision & 7.43 & -20.58 & 6.09$\%$ & 1 & 92 & 822 & 500 & 111 & 4 \\\shline

\rowcolor{lightgray}
\multicolumn{10}{c}{\emph{Open-Source MLLMs}}  \\ \shline
Qwen2.5-VL-72B-Instruct & 8.33 & -5.82 & 13.2$\%$ & 6 & 196 & 984 & 302 & 42 & 0 \\
InternVL2.5-78B-MPO & 8.06 & -12.55 & 8.76$\%$ & 6 & 128 & 917 & 434 & 45 & 0 \\
InternVL2.5-8B-MPO	&7.65 & -15.10 & 10.33$\%$ & 0 & 158 & 843 & 438 & 91 & 0 \\
InternVL2.5-78B	&7.91 & -16.43 & 7.25$\%$ & 4 & 107 & 863 & 494 & 62 & 0 \\
Qwen2-VL-72B-instruct&	7.87 & -22.45 & 4.64$\%$ & 0 & 71 & 764 & 632 & 63 & 0 \\
InternVL2.5-8B&	7.38 & -25.42 & 5.62$\%$ & 2 & 84 & 699 & 624 & 121 & 0 \\
Qwen2.5-VL-7B-Instruct&	7.55 & -29.80 & 4.25$\%$ & 0 & 65 & 620 & 713 & 132 & 0 \\
MiniCPM-o-2.6 &7.49 & -34.77 & 2.29$\%$ & 2 & 33 & 545 & 799 & 151 & 0 \\
DeepSeek-VL2&	7.24 & -38.52 & 1.77$\%$ & 0 & 27 & 504 & 791 & 207 & 1 \\
LLaVA-OneVision-72B	&7.16 & -39.87 & 1.72$\%$ & 0 & 26 & 448 & 842 & 194 & 20 \\
LLaVA-OneVision-7B	&6.75 & -43.49 & 1.96$\%$ & 1 & 29 & 411 & 816 & 273 & 0 \\
Qwen2-VL-7B-instruct	&7.12 & -43.76 & 1.57$\%$ & 0 & 24 & 402 & 845 & 259 & 0 \\\shline
\end{tabular}
}
\caption{\textbf{Win Rate Result of MLLMs on Creation-MMBench.} WR, MB, MW stands for Win Rate, Much Better and Much Worse}
\label{tab:win_rate_table}
\vspace{-4mm}
\end{table*}

%% file: tables/description_comparison_table.tex
\begin{table}[t]
\centering
\tablestyle{4pt}{1.1}
\resizebox{\linewidth}{!}{%
\begin{tabular}{l|cc|cc}
\shline
\multirow{2}{*}{\bf LLM} & \multicolumn{2}{c|}{\bf Generic} & \multicolumn{2}{c}{\bf Query-Specific} \\ \scline{2-5}
& \textbf {VFS} & \textbf {Reward}  & \textbf {VFS} & \textbf {Reward}  \\ \shline
GPT-4o-1120 & 8.71 & 6.96 &  8.88 & 3.33  \\ 
Qwen2.5-72B-Instruct & 8.55 & 0.82 & 8.82 & 4.80  \\ 
InternLM2.5-7B-Chat &	7.83 & -22.19 & 8.33 & -15.29\\ \shline

\end{tabular}
}

\caption{\textbf{Comparison on Generic Descriptions and Query-Specific Descriptions on Creation-MMBench-TO.} }
\label{tab: generic-query_specific comparison}
\end{table}

%% file: tables/human_judge_2.tex
\begin{table*}[t]
\centering
\tablestyle{10pt}{1.2}
\resizebox{\textwidth}{!}{%
\begin{tabular}{c|c|c|c|c|c|c|c|c|c|c}
\shline
\multirow{2}{*}{\bf Judging Method} & \multirow{2}{*}{\bf Judging Model/Human} & \multirow{2}{*}{\bf MLLM}  
& \multicolumn{4}{c|}{\bf Dual Evaluation} & \multicolumn{4}{c}{\bf Non-Dual Evaluation} \\
\cline{4-11}
& & & \multicolumn{2}{c|}{\textbf{MAE}$\downarrow$} & \multicolumn{2}{c|}{\textbf{Consistency}$\uparrow$} & \multicolumn{2}{c|}{\textbf{MAE}$\downarrow$} & \multicolumn{2}{c}{\textbf{Consistency}$\uparrow$} \\
\shline
\multirow{12}{*}{LLM-as-a-judge} 
& \multirow{3}{*}{Gemini-2.0-Pro} 
  & Gemini-1.5-pro-002 & 0.67 & \multirow{3}{*}{0.62} & 83.17 & \multirow{3}{*}{84.16} & 0.75 & \multirow{3}{*}{0.69} & 77.23 & \multirow{3}{*}{79.21} \\
& & Qwen2-VL-72B & 0.59 & & 84.16 & & 0.65 & & 78.22 & \\
& & MiniCPM-o-2.6 & 0.61 & & 85.15 & & 0.67 & & 82.18 & \\
\cline{2-11}
& \multirow{3}{*}{GPT-4o-mini} 
  & Gemini-1.5-pro-002 & 0.67 & \multirow{3}{*}{0.59} & 83.17 & \multirow{3}{*}{86.23} & 0.79 & \multirow{3}{*}{0.71} & 74.26 & \multirow{3}{*}{77.38} \\
& & Qwen2-VL-72B & 0.59 & & 85.29 & & 0.67 & & 76.47 & \\
& & MiniCPM-o-2.6 & 0.52 & & 90.20 & & 0.66 & & 81.37 & \\
\cline{2-11}
& \multirow{3}{*}{Claude-3.5-Sonnet} 
  & Gemini-1.5-pro-002 & 0.63 & \multirow{3}{*}{0.52} & 89.11 & \multirow{3}{*}{91.80} & 0.73 & \multirow{3}{*}{0.63} & 78.22 & \multirow{3}{*}{81.97} \\
& & Qwen2-VL-72B & 0.46 & & 94.12 & & 0.58 & & 82.35 & \\
& & MiniCPM-o-2.6 & 0.46 & & 92.16 & & 0.58 & & 85.29 & \\
\cline{2-11}
& \multirow{3}{*}{GPT-4o} 
  & Gemini-1.5-pro-002 & 0.56 & \multirow{3}{*}{0.51} & 93.07 & \multirow{3}{*}{91.48} & 0.56 & \multirow{3}{*}{0.56} & 90.10 & \multirow{3}{*}{87.54} \\
& & Qwen2-VL-72B & 0.46 & & 92.16 & & 0.54 & & 87.25 & \\
& & MiniCPM-o-2.6 & 0.51 & & 89.22 & & 0.58 & & 85.29 & \\
\shline
\multirow{12}{*}{MLLM-as-a-judge} 
& \multirow{3}{*}{Gemini-2.0-Pro} 
  & Gemini-1.5-pro-002 & 0.65 & \multirow{3}{*}{0.59} & 82.83 & \multirow{3}{*}{86.67} & 0.78 & \multirow{3}{*}{0.72} & 74.75 & \multirow{3}{*}{78.67} \\
& & Qwen2-VL-72B & 0.51 & & 91.00 & & 0.67 & & 80.00 & \\
& & MiniCPM-o-2.6 & 0.61 & & 86.14 & & 0.69 & & 81.19 & \\
\cline{2-11}
& \multirow{3}{*}{GPT-4o-mini} 
  & Gemini-1.5-pro-002 & 0.64 & \multirow{3}{*}{0.55} & 84.16 & \multirow{3}{*}{89.51} & 0.71 & \multirow{3}{*}{0.66} & 76.24 & \multirow{3}{*}{80.33} \\
& & Qwen2-VL-72B & 0.53 & & 93.14 & & 0.65 & & 82.35 & \\
& & MiniCPM-o-2.6 & 0.49 & & 91.18 & & 0.61 & & 82.35 & \\
\cline{2-11}
& \multirow{3}{*}{Claude-3.5-Sonnet} 
  & Gemini-1.5-pro-002 & 0.56 & \multirow{3}{*}{0.50} & 89.90 & \multirow{3}{*}{90.60} & 0.61 & \multirow{3}{*}{0.59} & 83.84 & \multirow{3}{*}{85.23} \\
& & Qwen2-VL-72B & 0.46 & & 92.00 & & 0.59 & & 85.00 & \\
& & MiniCPM-o-2.6 & 0.47 & & 89.90 & & 0.57 & & 86.87 & \\
\cline{2-11}
& \multirow{3}{*}{\textbf{GPT-4o}} 
  & Gemini-1.5-pro-002 & 0.53 & \multirow{3}{*}{\textbf{0.50}} & 92.08 & \multirow{3}{*}{\textbf{92.13}} & 0.57 & \multirow{3}{*}{\textbf{0.54}} & 89.11 & \multirow{3}{*}{\textbf{88.85}} \\
& & Qwen2-VL-72B & 0.42 & & 96.08 & & 0.46 & & 91.18 & \\
& & MiniCPM-o-2.6 & 0.53 & & 88.24 & & 0.59 & & 86.27 & \\

\shline
\multirow{12}{*}{Human-as-a-judge} 
& \multirow{3}{*}{H1} 
  & Gemini-1.5-pro-002 & / & \multirow{3}{*}{/} & / & \multirow{3}{*}{/} & 0.65 & \multirow{3}{*}{0.64} & 84.16 & \multirow{3}{*}{87.21} \\
& & Qwen2-VL-72B & / & & / & & 0.60 & & 90.20 & \\
& & MiniCPM-o-2.6 & / & & / & & 0.66 & & 87.25 & \\
\cline{2-11}
& \multirow{3}{*}{H2} 
  & Gemini-1.5-pro-002 & / & \multirow{3}{*}{/} & / & \multirow{3}{*}{/} & 0.82 & \multirow{3}{*}{0.75} & 74.26 & \multirow{3}{*}{78.69} \\
& & Qwen2-VL-72B & / & & / & & 0.72 & & 82.35 & \\
& & MiniCPM-o-2.6 & / & & / & & 0.73 & & 79.41 & \\
\cline{2-11}
& \multirow{3}{*}{H3} 
  & Gemini-1.5-pro-002 & / & \multirow{3}{*}{/} & / & \multirow{3}{*}{/} & 0.74 & \multirow{3}{*}{0.68} & 76.24 & \multirow{3}{*}{82.30} \\
& & Qwen2-VL-72B & / & & / & & 0.62 & & 80.39 & \\
& & MiniCPM-o-2.6 & / & & / & & 0.72 & & 90.20 & \\
\cline{2-11}
& \multirow{3}{*}{H4} 
  & Gemini-1.5-pro-002 & / & \multirow{3}{*}{/} & / & \multirow{3}{*}{/} & 0.64 & \multirow{3}{*}{0.63} & 87.13 & \multirow{3}{*}{87.87} \\
& & Qwen2-VL-72B & / & & / & & 0.61 & & 89.22 & \\
& & MiniCPM-o-2.6 & / & & / & & 0.65 & & 87.25 & \\

\shline
\end{tabular}
}
\caption{\textbf{The Results of Model-Human Alignment and Human-Human Alignment.}}
\label{tab:model-human-alignment}
\end{table*}

%% file: tables/category_case.tex
\begin{figure*}[t]
\centering
\includegraphics[width=\linewidth]{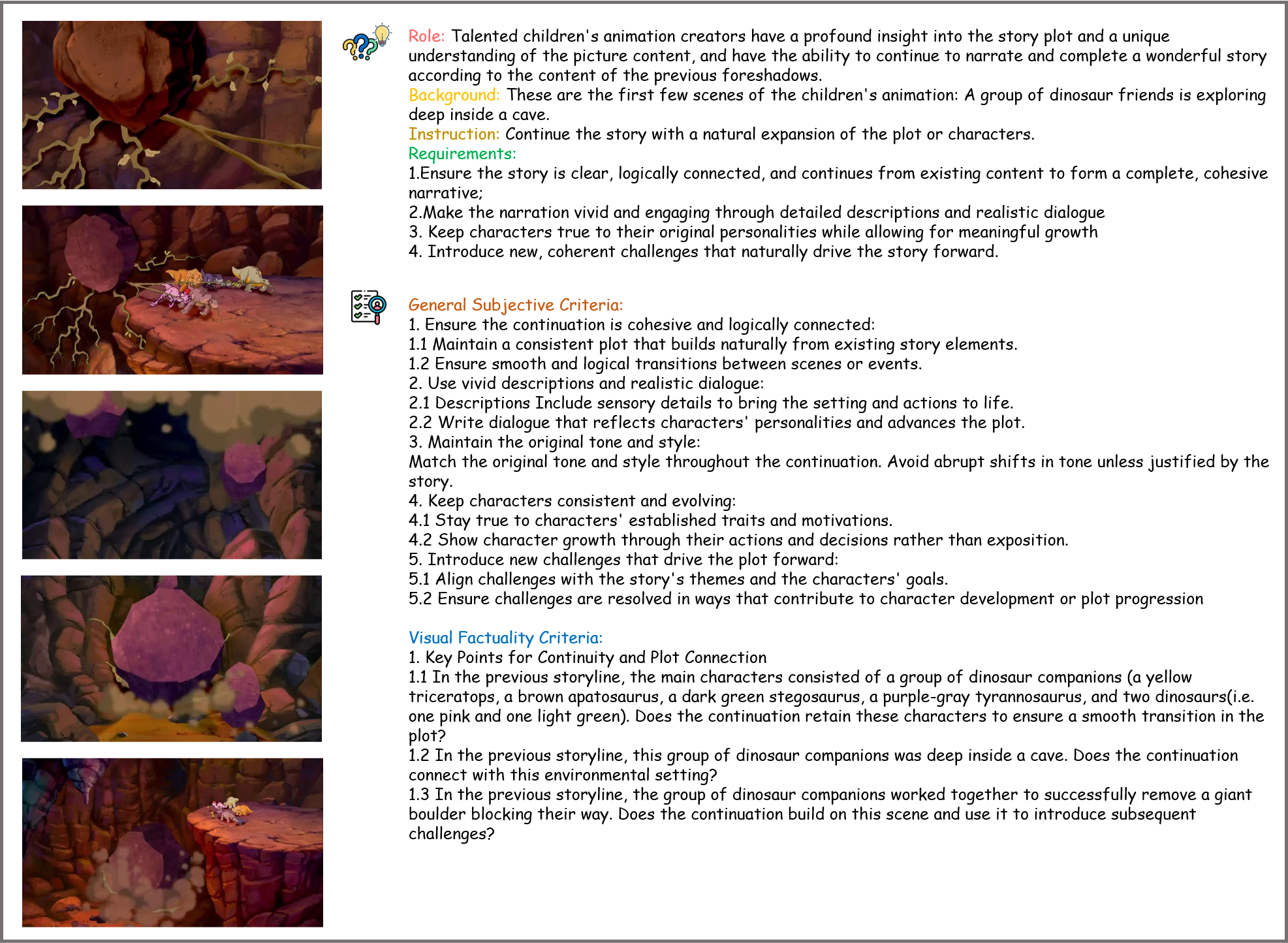}
\caption{\textbf{Example Case of Literary Writing, from Task story continue.}}
\label{fig:LW_story_continue}
\end{figure*}

\begin{figure*}[t]
\centering
\includegraphics[width=\linewidth]{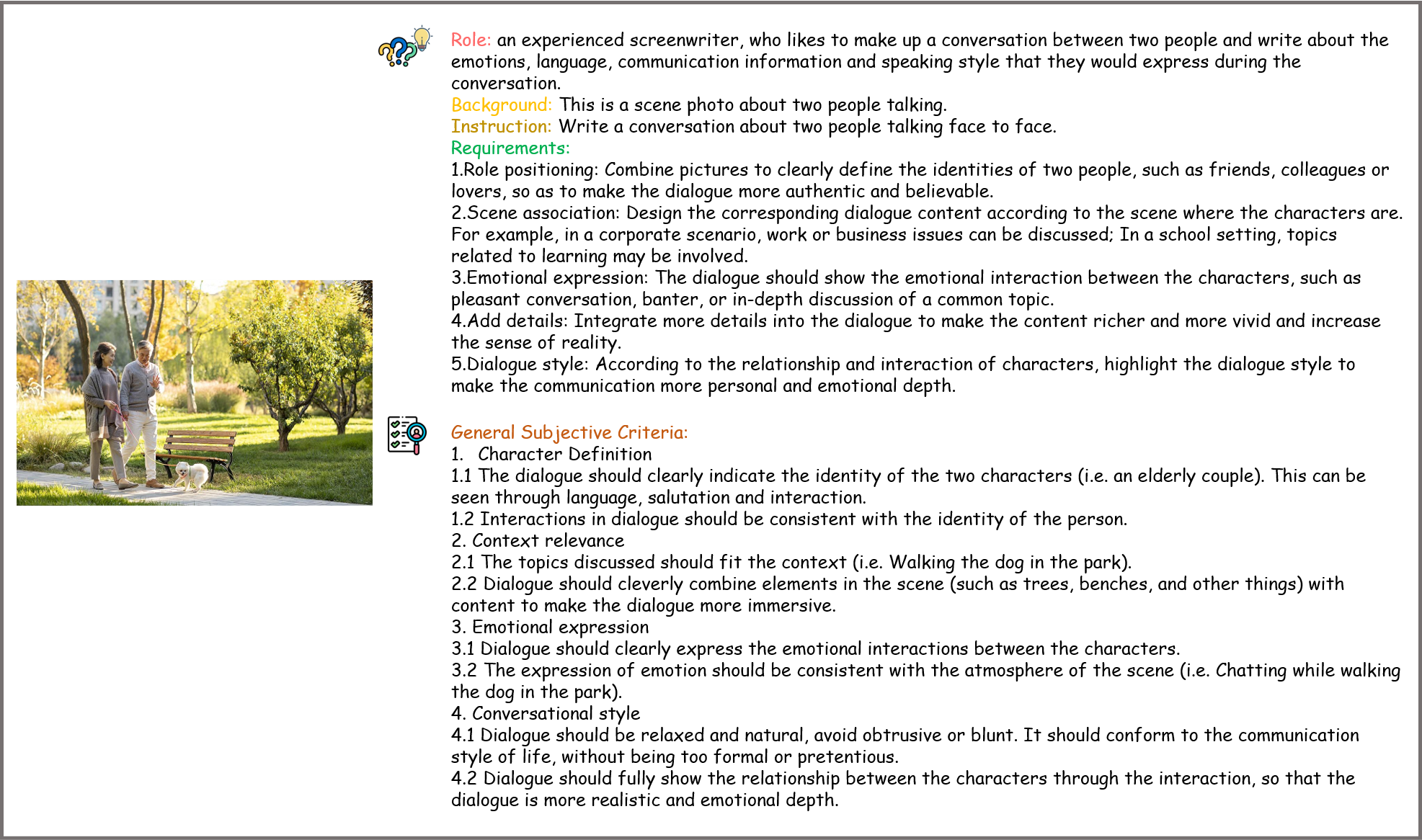}
\caption{\textbf{Example Case of Literary Writing, from Task daily conversation creation.}}
\label{fig:LW_daily_conversation_creation}
\end{figure*}

\begin{figure*}[t]
\centering
\includegraphics[width=\linewidth]{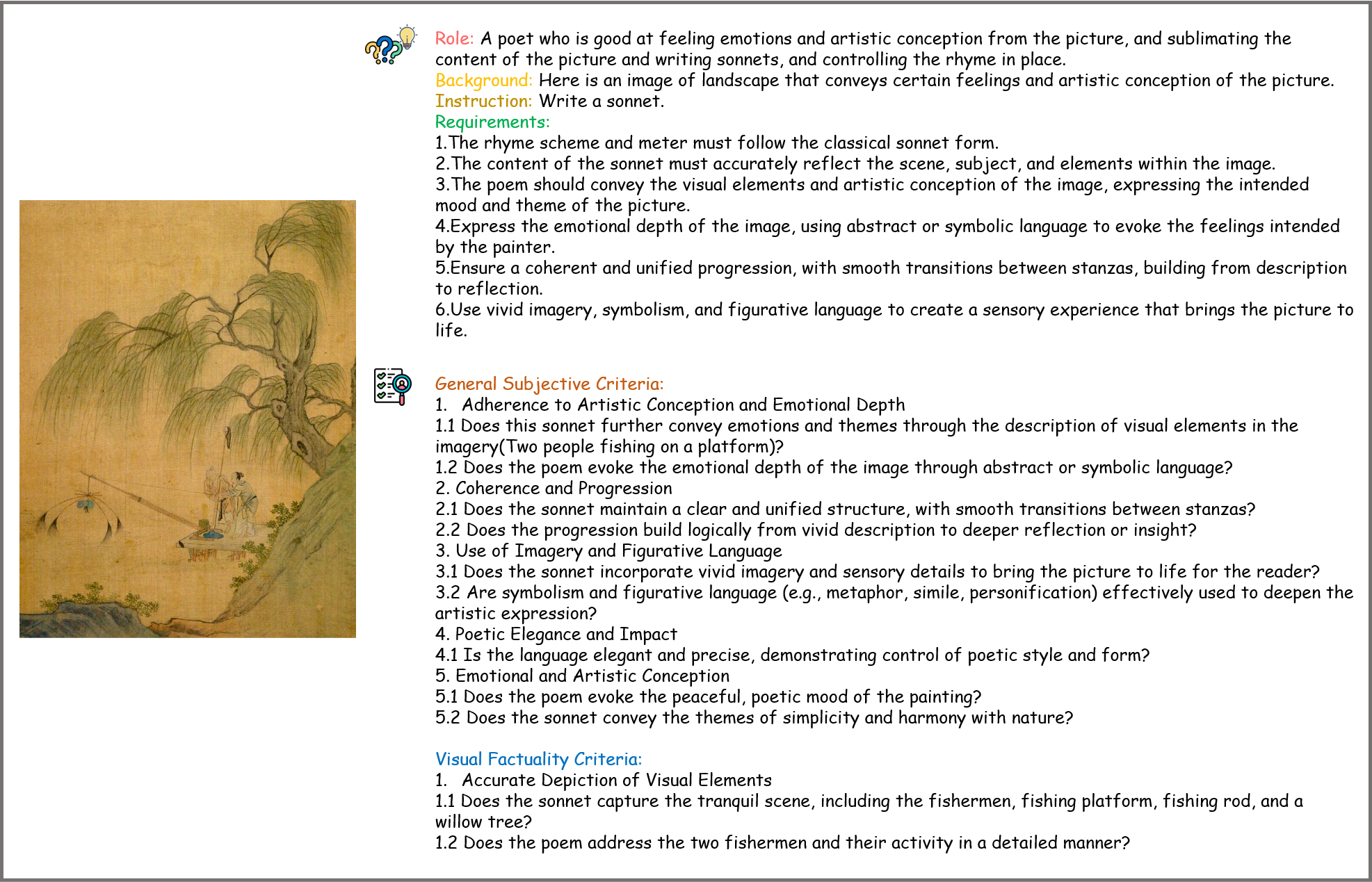}
\caption{\textbf{Example Case of Literary Writing, from Task landscape to poem.}}
\label{fig:LW_landscape_to_poem}
\end{figure*}

\begin{figure*}[t]
\centering
\includegraphics[width=\linewidth]{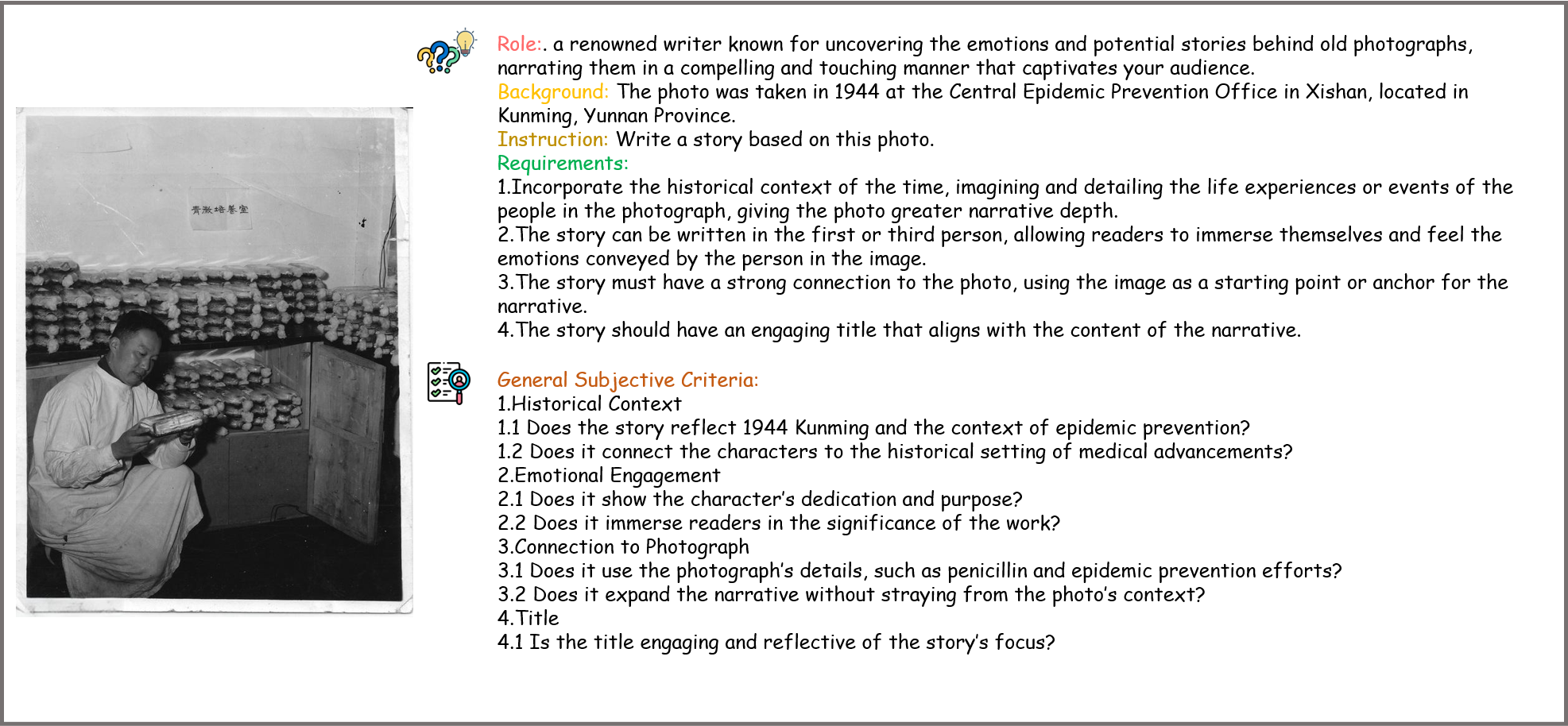}
\caption{\textbf{Example Case of Literary Writing, from Task historical story creation.}}
\label{fig:LW_historical_story_creation}
\end{figure*}

\begin{figure*}[t]
\centering
\includegraphics[width=\linewidth]{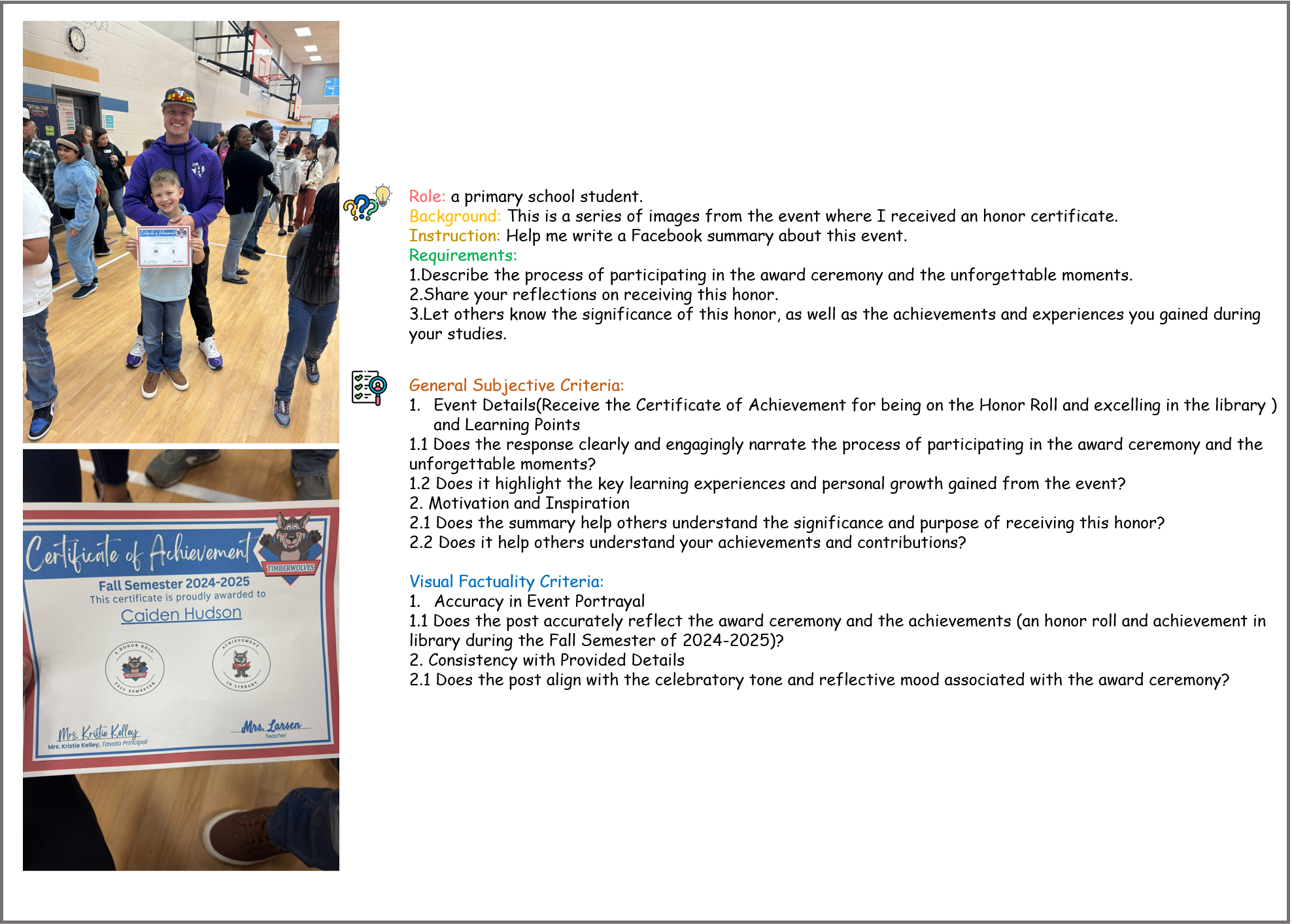}
\caption{\textbf{Example Case of Common Functional Writing, from Task daily achievement show off.}}
\label{fig:CFW_daily_achievement_show_off}
\end{figure*}

\begin{figure*}[t]
\centering
\includegraphics[width=\linewidth]{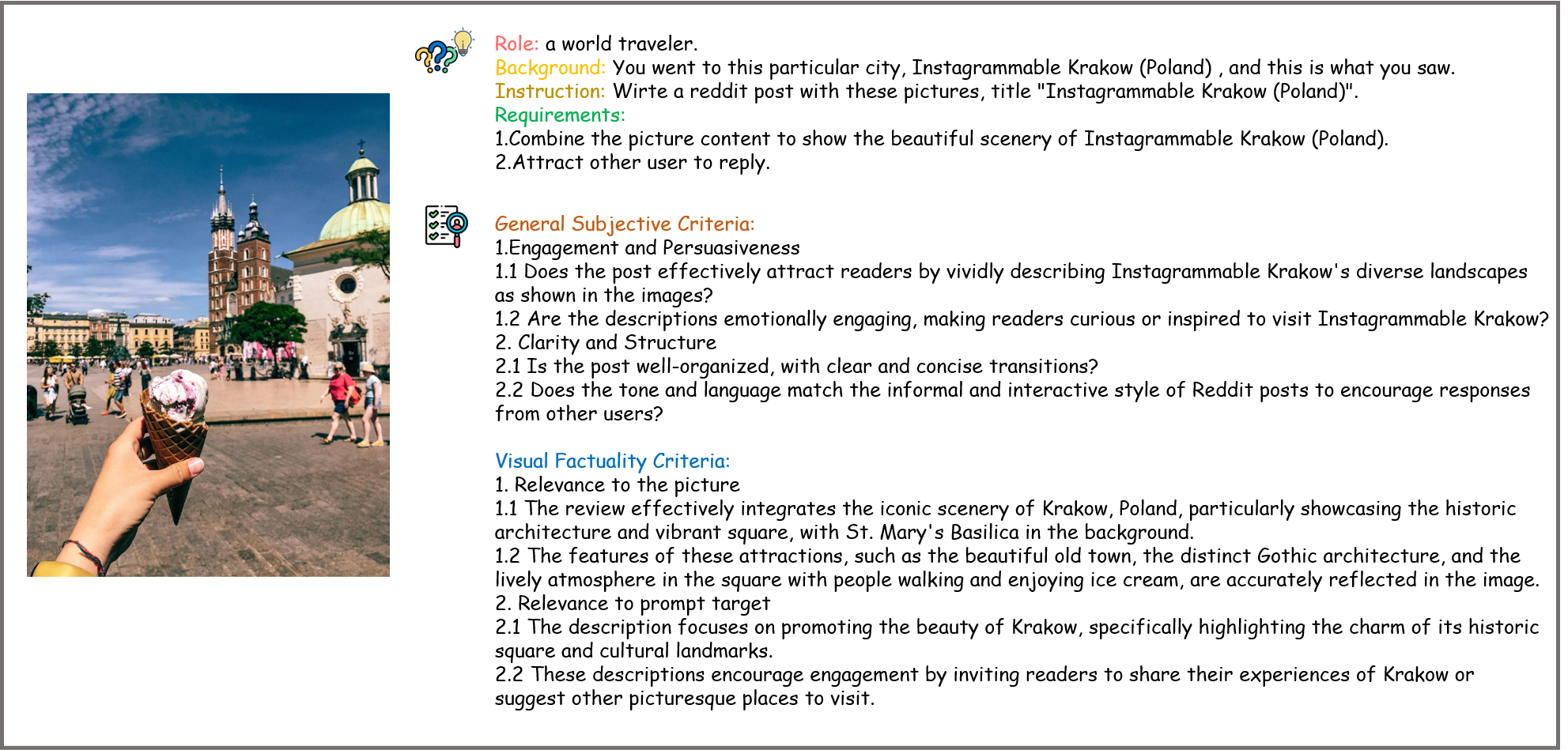}
\caption{\textbf{Example Case of Common Functional Writing, from Task social media travel content.}}
\label{fig:CFW_social_media_travel_content}
\end{figure*}

\begin{figure*}[t]
\centering
\includegraphics[width=\linewidth]{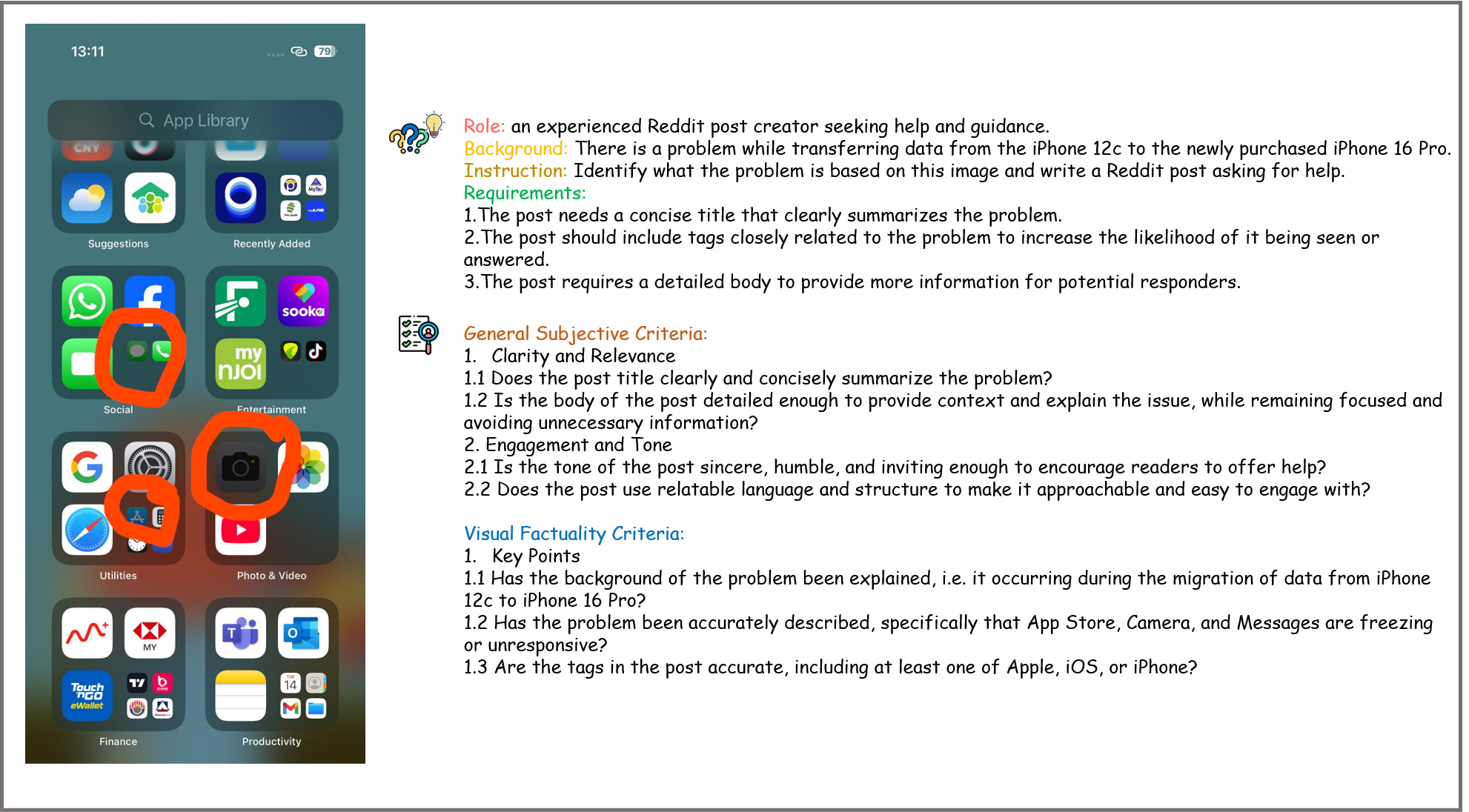}
\caption{\textbf{Example Case of Common Functional Writing, from Task daily affairs inquiries.}}
\label{fig:CFW_daily_affairs_inquiries}
\end{figure*}

\begin{figure*}[t]
\centering
\includegraphics[width=\linewidth]{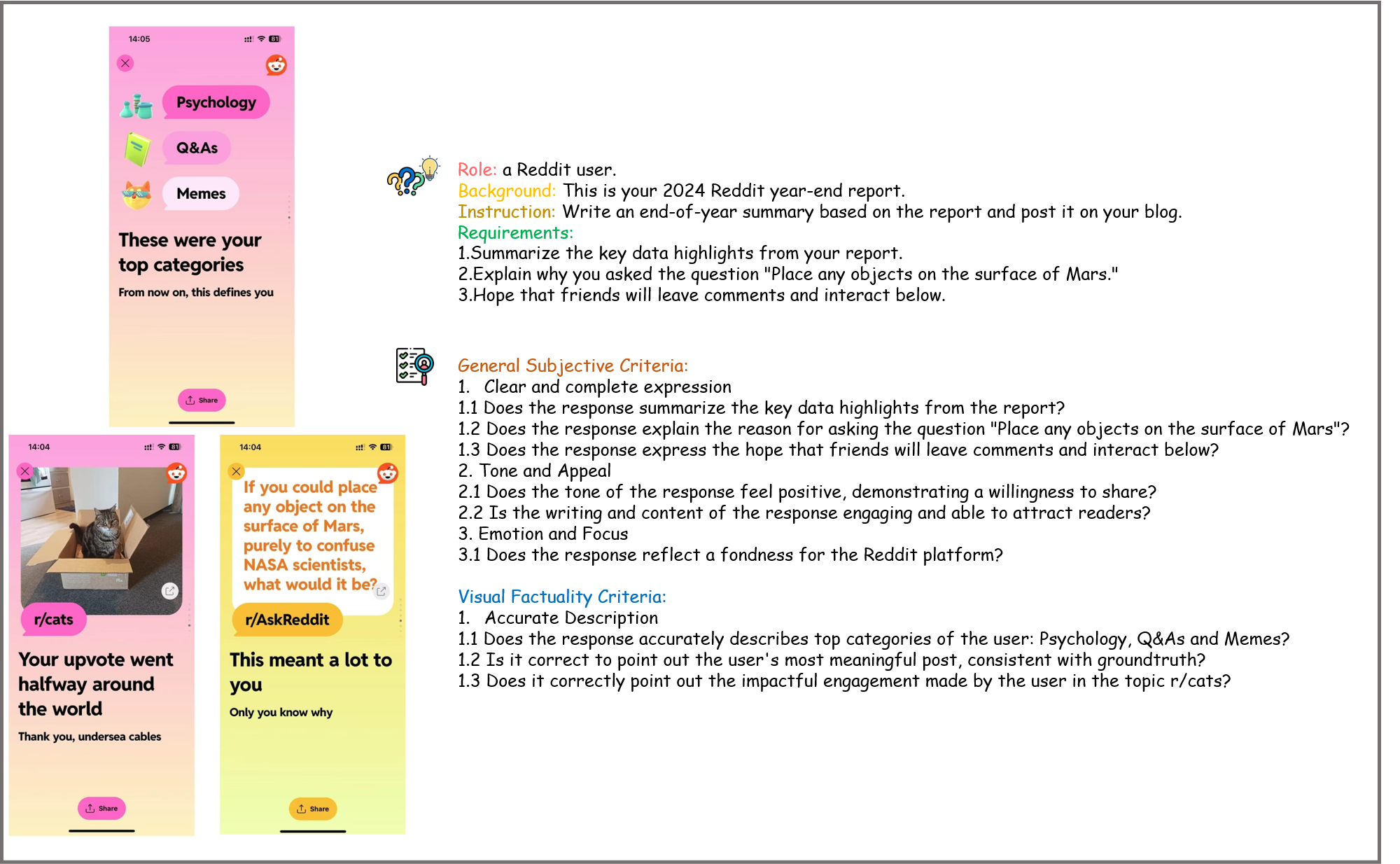}
\caption{\textbf{Example Case of Common Functional Writing, from Task personal event summaries.}}
\label{fig:CFW_personal_event_summaries}
\end{figure*}

\begin{figure*}[t]
\centering
\includegraphics[width=\linewidth]{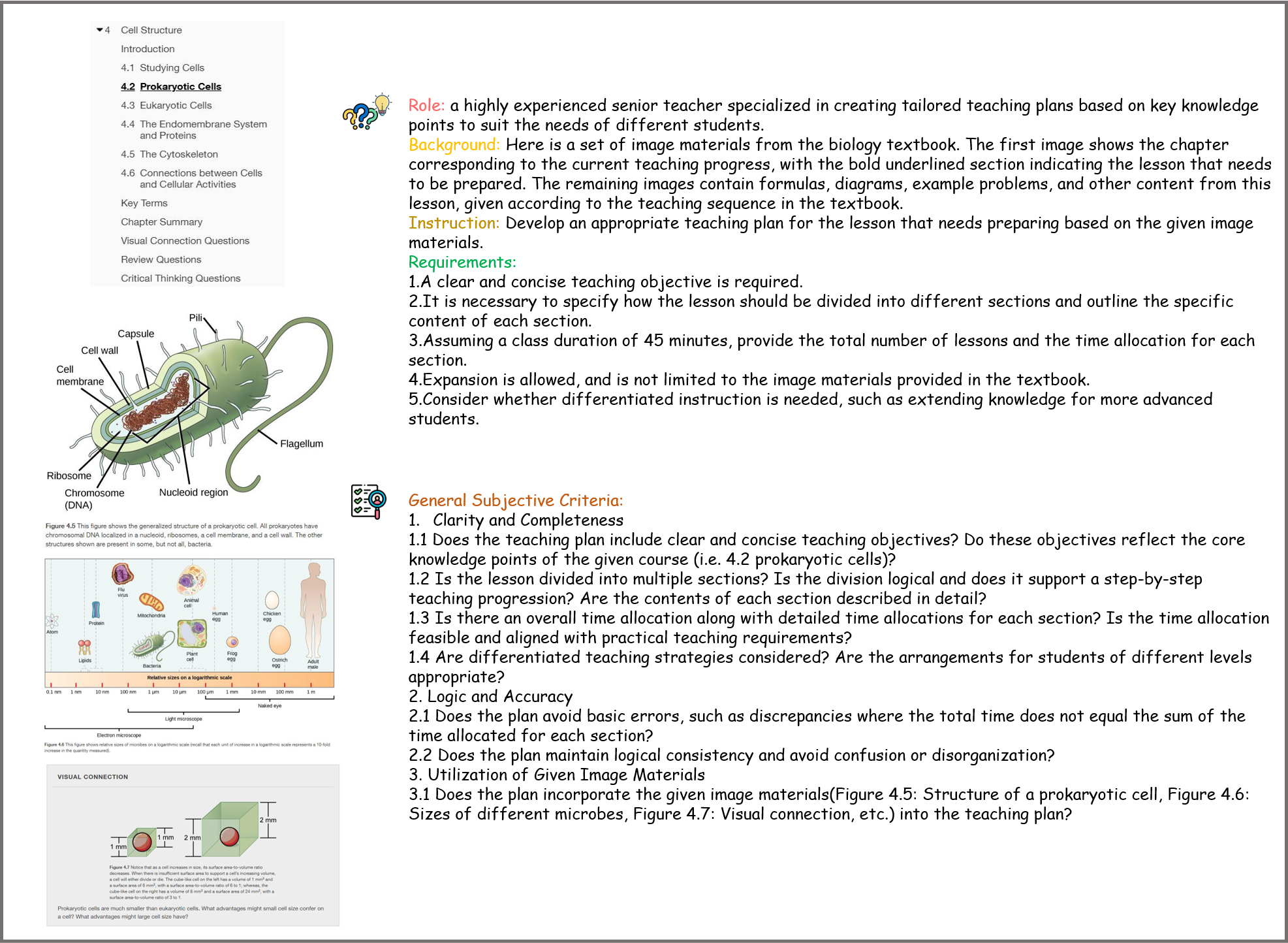}
\caption{\textbf{Example Case of Professional Functional Writing, from Task teaching plan.}}
\label{fig:PFW_teaching_plan}
\end{figure*}

\begin{figure*}[t]
\centering
\includegraphics[width=\linewidth]{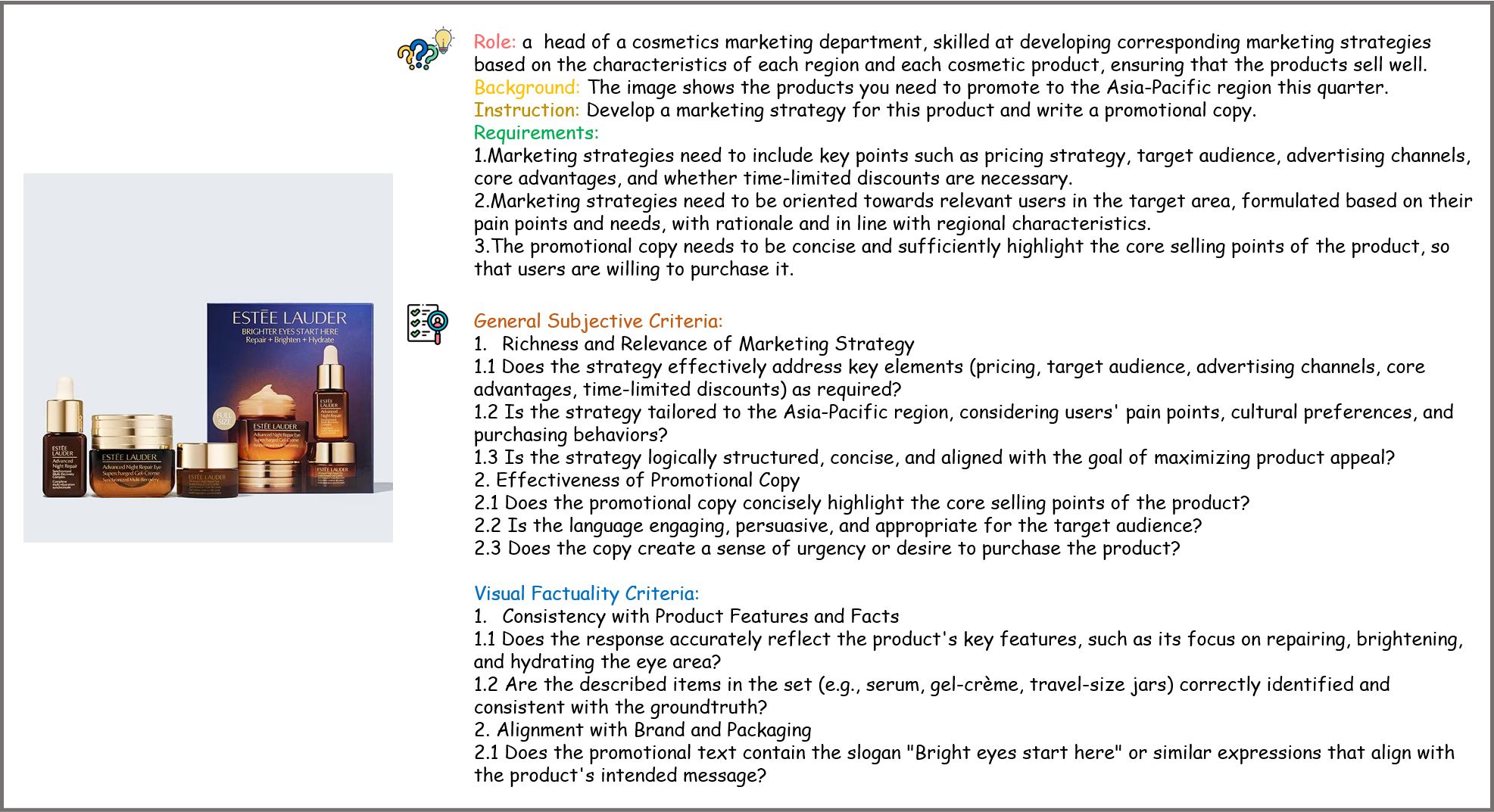}
\caption{\textbf{Example Case of Professional Functional Writing, from Task product marketing strategy.}}
\label{fig:PFW_product_marketing_strategy}
\end{figure*}

\begin{figure*}[t]
\centering
\includegraphics[width=\linewidth]{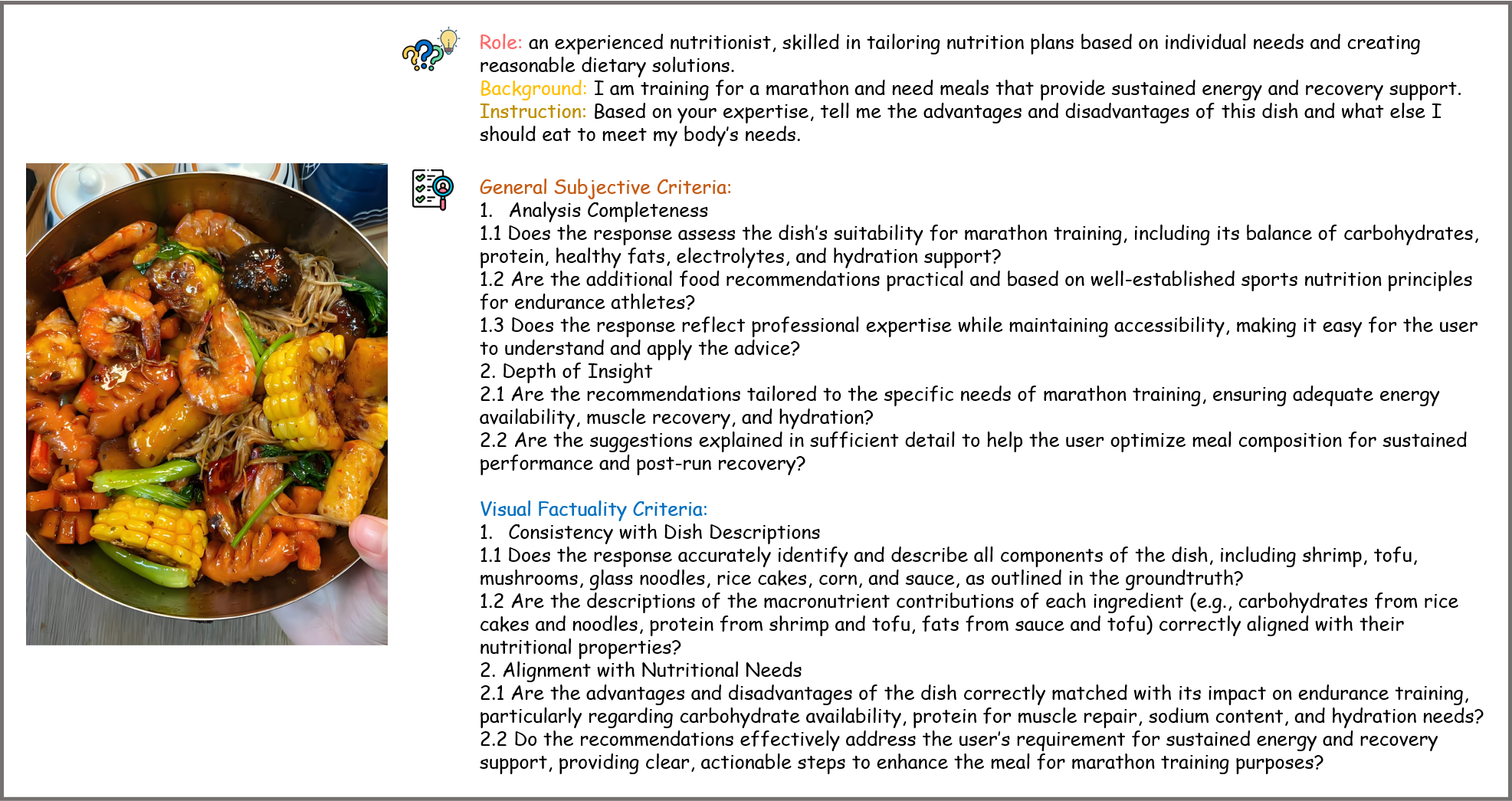}
\caption{\textbf{Example Case of Professional Functional Writing, from Task nutritional formulation of recipe.}}
\label{fig:PFW_nutritional_formulation_of_recipe}
\end{figure*}

\begin{figure*}[t]
\centering
\includegraphics[width=\linewidth]{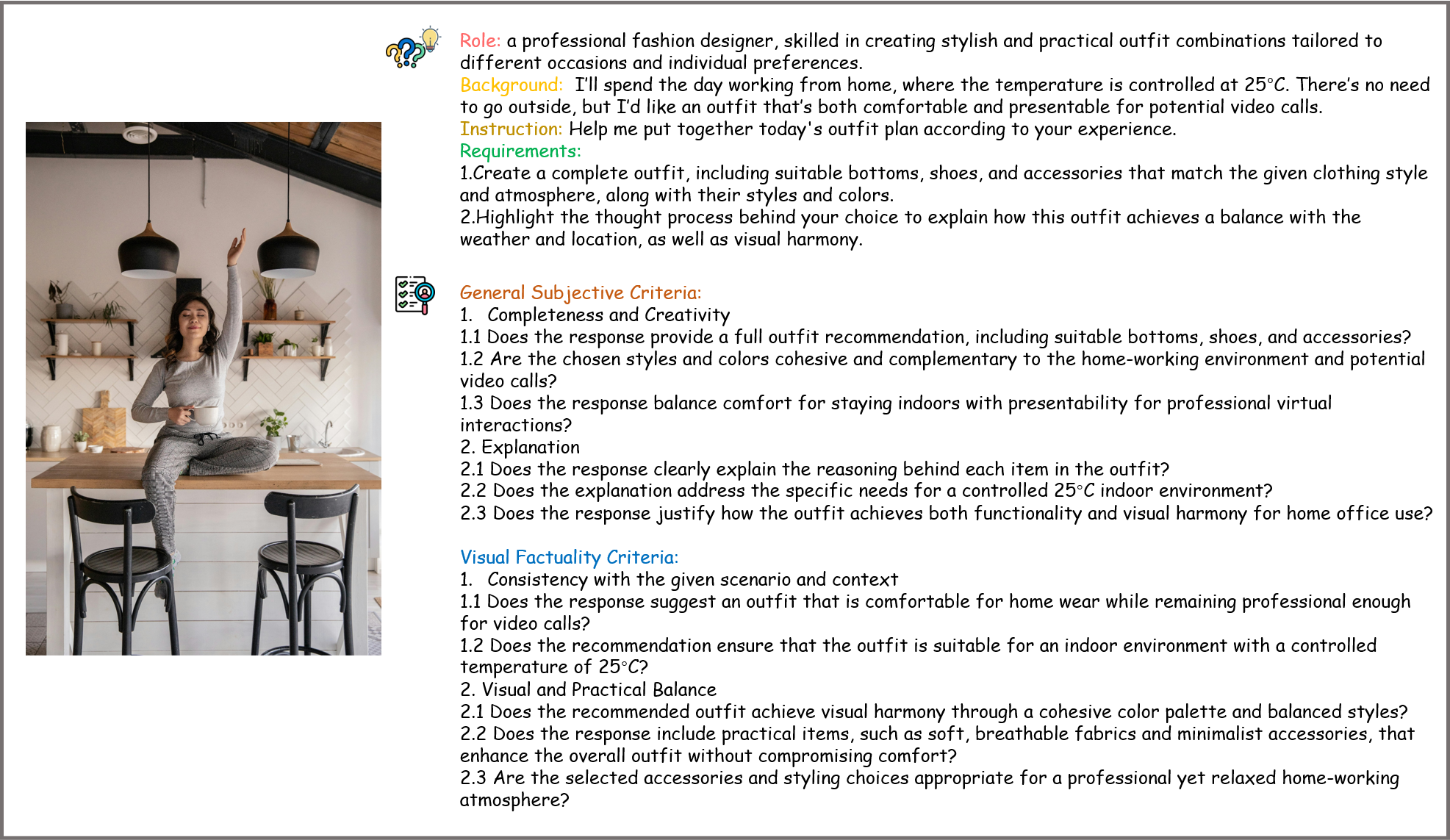}
\caption{\textbf{Example Case of Professional Functional Writing, from Task clothing match design.}}
\label{fig:PFW_clothing_match_design}
\end{figure*}

\begin{figure*}[t]
\centering
\includegraphics[width=\linewidth]{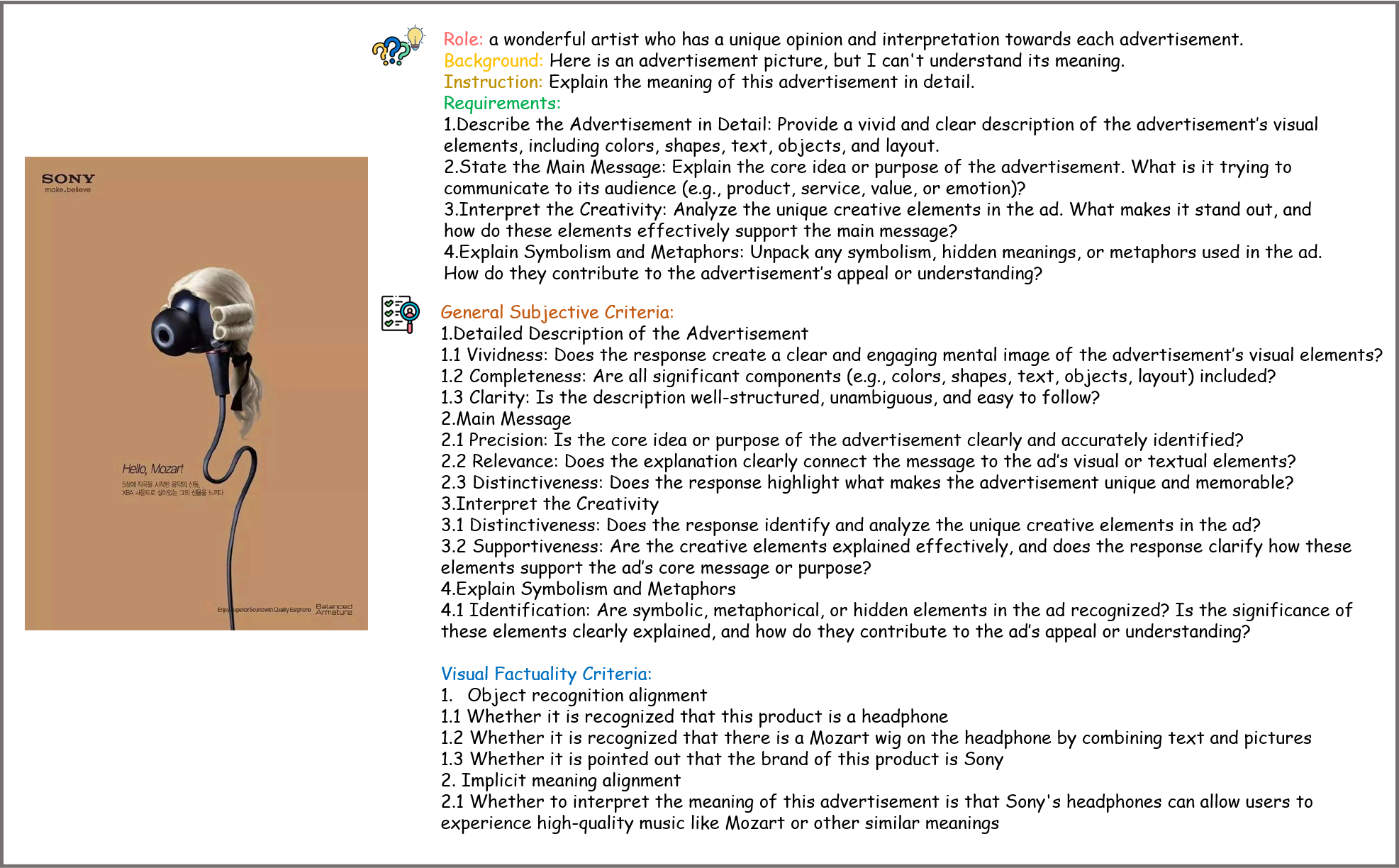}
\caption{\textbf{Example Case of Creative Multimodal Understanding, from Task advertisement explanation.}}
\label{fig:CMU_advertisement_explanation}
\end{figure*}

\begin{figure*}[t]
\centering
\includegraphics[width=\linewidth]{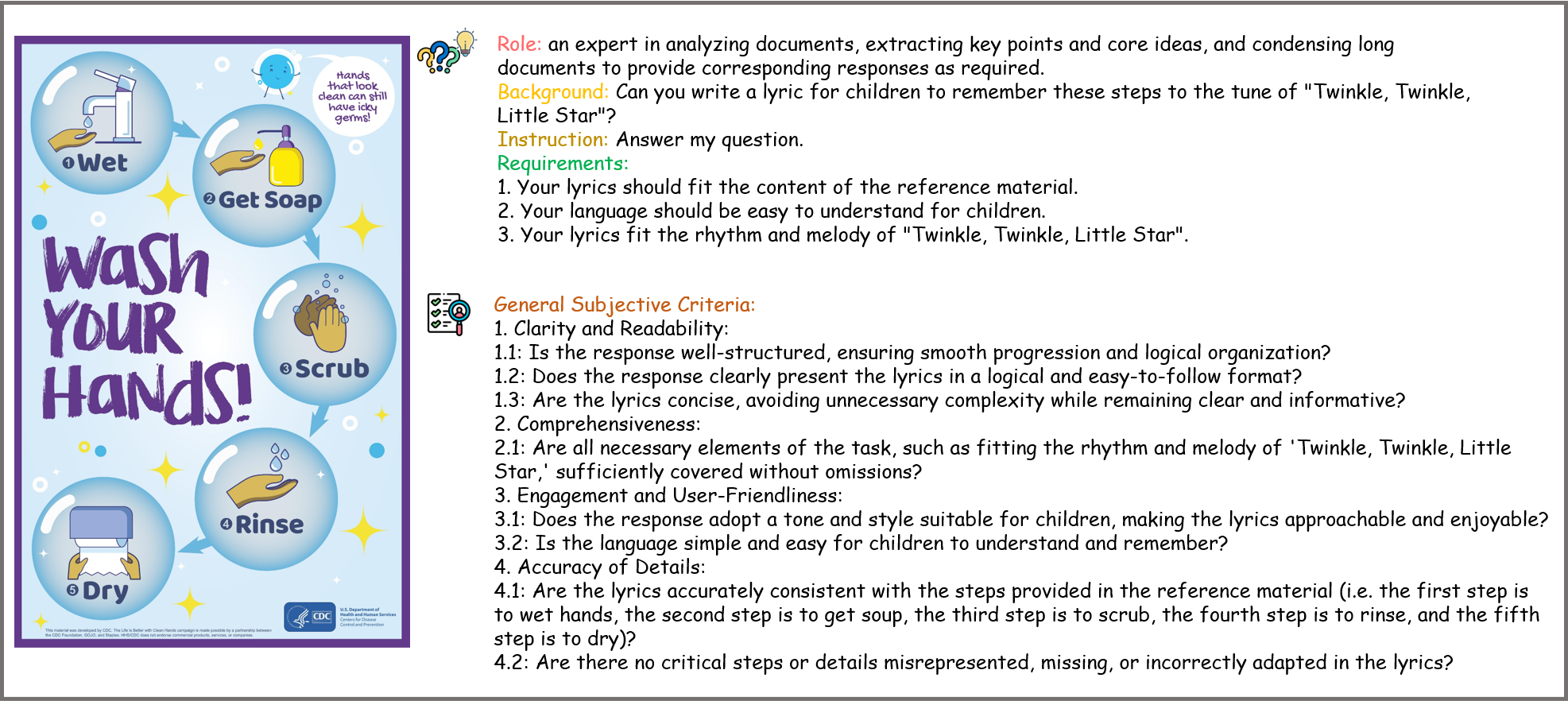}
\caption{\textbf{Example Case of Creative Multimodal Understanding, from Task document understanding.}}
\label{fig:CMU_document_understanding}
\end{figure*}

\begin{figure*}[t]
\centering
\includegraphics[width=\linewidth]{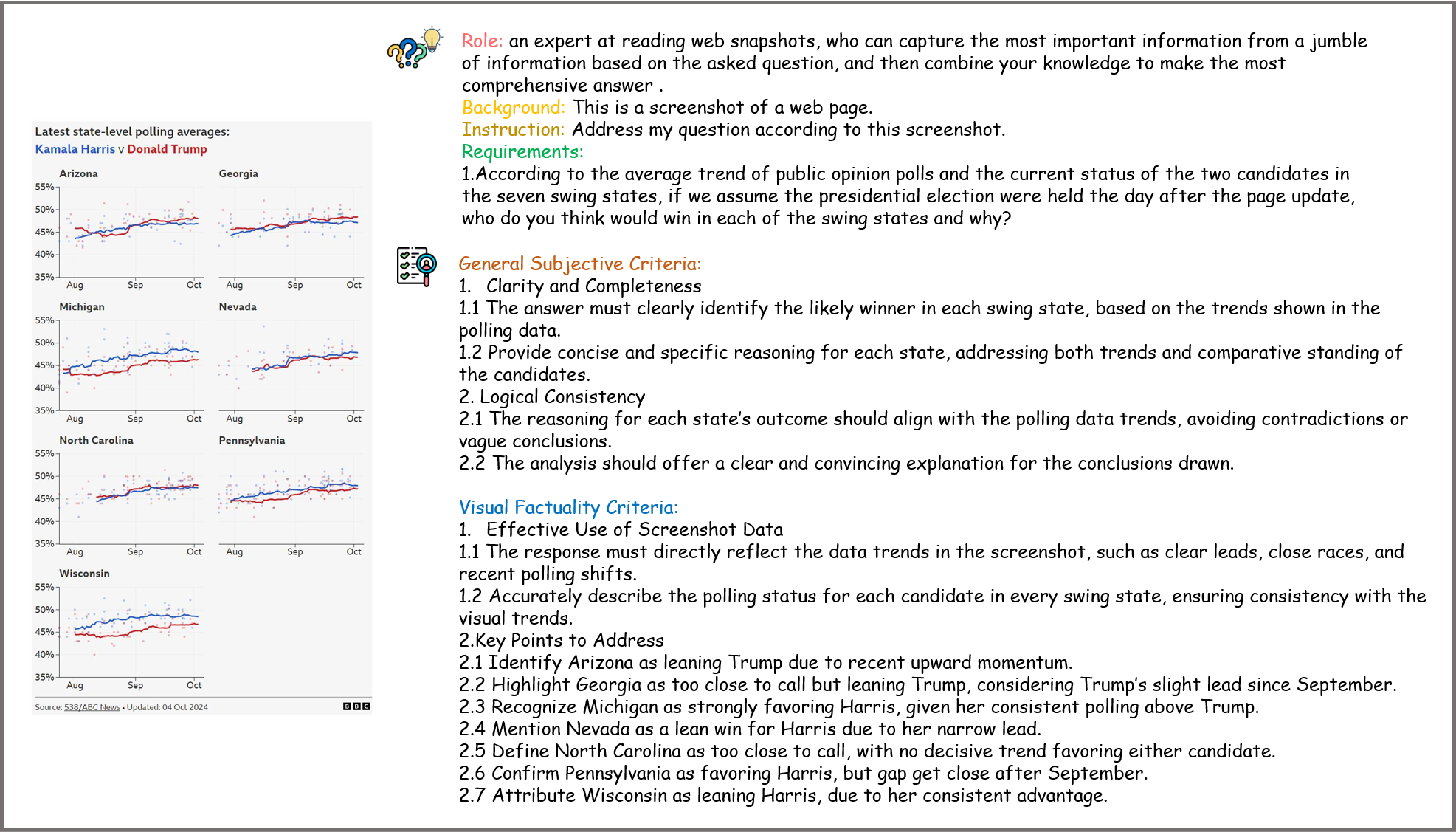}
\caption{\textbf{Example Case of Creative Multimodal Understanding, from Task snapshot analysis.}}
\label{fig:CMU_snapshot_analysis}
\end{figure*}

\begin{figure*}[t]
\centering
\includegraphics[width=\linewidth]{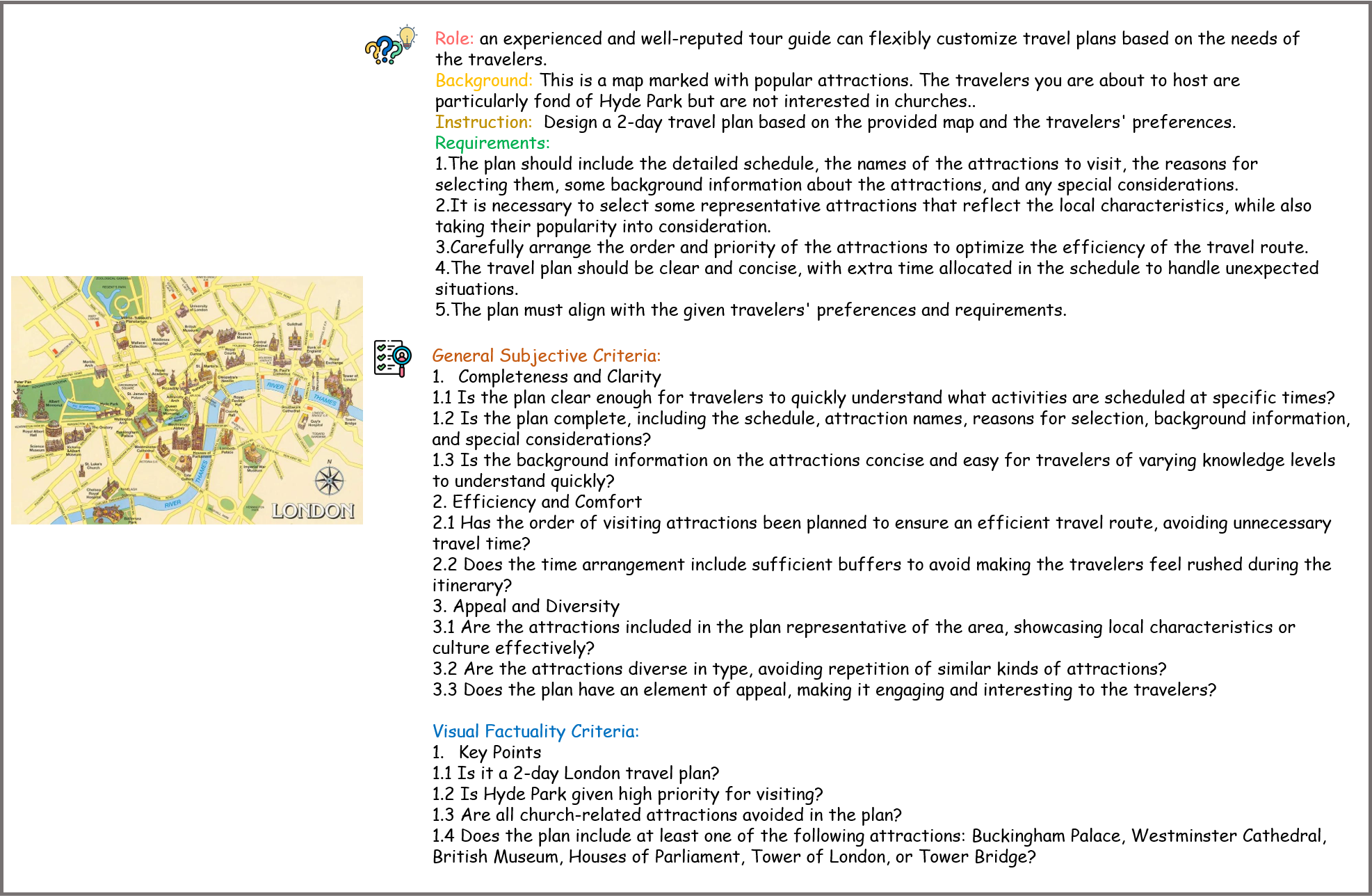}
\caption{\textbf{Example Case of Creative Multimodal Understanding, from Task travel itinerary planning and recommendations.}}
\label{fig:CMU_travel_itinerary_planning_and_recommendations}
\end{figure*}

%% file: tables/prompt_template.tex
\begin{figure*}[!ht]
    \centering
    
    \begin{minipage}{0.9\textwidth}
        \centering
        \begin{AIbox}{Query}
        {   
            \textbf{\textcolor{blue}{Creation-MMBench: }}
            
            Assume you are \textcolor{brown}{\textless Role\textgreater}
            
            \textcolor{brown}{\textless Background\textgreater}
            
            Please follow the requirements below to \textcolor{brown}{\textless Instruction\textgreater}.
            
            \textcolor{brown}{\textless Requirement\textgreater}

            \textbf{\textcolor{blue}{Creation-MMBench-TO: }}
            
            Assume you are \textcolor{brown}{\textless Role\textgreater}
            
            \textcolor{brown}{\textless Background\textgreater}
            
            Please follow the requirements below to \textcolor{brown}{\textless Instruction\textgreater}.
            
            \textcolor{brown}{\textless Requirement\textgreater}

            This question does not provide images, only descriptions of images by a large language model. Please answer based on the descriptions.
        
            Description of the image: \textcolor{brown}{\textless Image Description\textgreater}
        }
        \end{AIbox}
        \caption{\textbf{Query Template of Creation-MMBench and Creation-MMBench-TO}}
        \label{fig:Question_prompt_cmb}
    \end{minipage}
    

    
    \vspace{5mm} 

    \begin{minipage}{0.9\textwidth}
        \centering
        \begin{AIbox}{Descrption}
        {
            \textbf{\textcolor{blue}{Generic Instruction: }} Please carefully describe the content of each incoming image, starting with the number of images. For each image, first provide a general introduction to the content, then describe the image type, characters and objects, scene and atmosphere, the relationships between people and objects, and any text on the image. 
            
            \textbf{\textcolor{blue}{Query-Specific Instruction: }}Please carefully describe the content of each incoming image, starting with the number of images. For each image, first provide a general introduction to the content, then describe the image type, characters and objects, scene and atmosphere, the relationships between people and objects, and any text on the image. Please pay special attention to the following aspects: \textcolor{brown}{\textless query-specific part\textgreater}.
        }
        \end{AIbox}
        \caption{\textbf{Generic Instruction \textit{vs.} Query-Specific Instruction of Image Description}}
        \label{fig:description_prompt}
    \end{minipage}

\end{figure*}

\begin{figure*}[!ht] 
        \begin{AIbox}{Few-Shot Prompt Template}
        {   
            Your task is to give a concise instruction about what basic elements are needed to be described based on the given question. Ensure that your instructions do not cover the raw question, options, or thought process of answering the question. \\
            \\
            \#\#Question\#\#: Assume you are an expert at parsing documents, extracting key points and core ideas from documents, and condensing long documents into one or two paragraphs to summarize them. I come from the environmental protection department. What conclusions can I draw from this report? Please provide key evidence from the document to support your answer, within 100 words. Please follow the requirements below to Answer my question.
            
            1. Your answer should fit the content of the reference material.
            
            2. Provide a concise answer with sufficient thinking, removing unnecessary details.
            
            \#\#Contents to observe\#\#: All text and related charts on the picture show the trend of changes \\
            \\
            \#\#Question\#\#: Assume you are a world traveler. You went to this particular country, Chile, and this is what you saw. Please follow the requirements below to write a Reddit post with these pictures, titled "Am I the only one who feels Chile is extremely underrated as a travel destination?"
            
            1. Combine the picture content to show the beautiful scenery of Chile.
            
            2. Attract other users to reply
            
            \#\#Contents to observe\#\#: The uniqueness of the main scenery in the picture and the emotions conveyed by the characters in the picture \\
            \\
            \#\#Question\#\#: Assume you are an experienced Reddit post creator seeking help and guidance. There is a problem while uninstalling a software from the iPhone 12. Please follow the requirements below to identify what the problem is based on this image and write a Reddit post asking for help.
            
            1. The post needs a concise title that clearly summarizes the problem
            
            2. The post should include tags closely related to the problem to increase the likelihood of it being seen or answered
            
            3. The post requires a detailed body to provide more information for potential responders
            
            \#\#Contents to observe\#\#: The situation on the interface and the main issues identified \\
            \\
            \#\#Question\#\#: Assume you are experienced UI/UX designer, skilled in analyzing and optimizing interface designs to enhance usability and visual appeal. Please follow the requirements below to analyze and propose optimization suggestions for the current UI design based on the provided image.
            
            1. The goal of the optimization is to improve the interface layout, user interaction flow, and overall visual aesthetics while ensuring a seamless user experience.
            
            2. After optimization, the application should have strong visual appeal, easy navigation, and provide users with an enjoyable experience, ultimately increasing user engagement and retention rates.
            
            \#\#Contents to observe\#\#: Interface layout and related information on the interface \\
            \\
            \#\#Question\#\#: \textcolor{brown}{\textless Question\textgreater}.
            
            \#\#Contents to observe\#\#: 
        }
        \end{AIbox}
        \caption{\textbf{The Prompt Template for the GPT-4o to Generate the "Query-Specific Part".}}
        \label{fig:few_shot_prompt}
\end{figure*}

\begin{figure*}[!ht] 
\begin{AIbox}{Subjective Judge}
{ 
\textcolor{brown}{\textless Image Content\textgreater} 

Please act as an impartial judge and evaluate the quality of the responses provided by two AI assistants to the user prompt below, considering both the provided criteria and the image.

Your task is to carefully assess each response based on how well it meets the evaluation criteria, incorporating the visual context from the image. The criteria should be the primary basis for your judgment, with the image serving to complement and inform your analysis.

Steps for Evaluation:

1.	Review Both Responses Independently:

    \quad Carefully analyze Assistant A’s and Assistant B’s responses with the criteria and the image. Do not assume any response is better just because it is listed first. Each response should be independently assessed based on the criteria and aided by images to help understand the context.

2.	Compare the Strengths and Weaknesses:

    \quad After evaluating each response independently, compare the two. Consider both the quality of the content and how closely it aligns with the criteria and image. Identify the strengths and weaknesses of each response, and highlight the key differences.

3.	Ensure Fairness:

    \quad To avoid positional bias, swap the positions of Assistant A and Assistant B after the first evaluation (i.e., make Assistant A become Assistant B and vice versa) and repeat the analysis and comparison. This ensures that each response is evaluated impartially under the same criteria.

4.	Provide a Conclusion Based on Both Evaluations:

    \quad After completing both evaluations (original and swapped positions), combine your analysis to provide a final verdict. If the responses are similar, with only minimal differences, your judgment should reflect that and indicate a tie.

Possible Verdict Options:

• If Assistant A is clearly better in both evaluations: [[A\verb|>>|B]]

• If Assistant A is slightly better in both evaluations: [[A\verb|>|B]]

• If both responses are nearly identical, showing minimal differences and no clear advantage: [[A\verb|=|B]]

• If Assistant B is slightly better in both evaluations: [[B\verb|>|A]]

• If Assistant B is clearly better in both evaluations: [[B\verb|>>|A]]

Instructions to the AI Assistants:

[INSTRUCTIONS]
\textcolor{brown}{\textless instructions\textgreater}
[END INSTRUCTIONS]

Assistant A Response:

[ASSISTANT A]
\textcolor{brown}{\textless Reference Answer\textgreater}
[END ASSISTANT A]

Evaluation Criteria:

[CRITERIA]
\textcolor{brown}{\textless Subjective Criteria\textgreater}
[END CRITERIA]

Assistant B Response:

[ASSISTANT B]
\textcolor{brown}{\textless Model Prediction\textgreater}
[END ASSISTANT B]

Output Format:

Your output should include:

1.	Evaluation of Assistant A’s Response: Provide a detailed qualitative evaluation, focusing on how well Assistant A’s response aligns with the criteria and the image.

2.	Evaluation of Assistant B’s Response: Provide a detailed qualitative evaluation, focusing on how well Assistant B’s response aligns with the criteria and the image.

3.	Final Verdict: After considering both evaluations, select one of the following verdicts and justify it based on your analysis:

Your output format should end like this:

Assistant A Evaluation: [qualitative comment]

Assistant B Evaluation: [qualitative comment]

Final Verdict is: [[VERDICT]]
}
\end{AIbox} 
\caption{\textbf{Subjective Judge Prompt Template of Creation-MMBench}}
\label{fig: sub_judge_creation}
\end{figure*}

\begin{figure*}[!ht] 
\begin{AIbox}{Visual Judge}
{ 

\textbf{\textcolor{blue}{With GroundTruth: }} 

Please act as an impartial judge and evaluate the Visual Factuality of the responses provided by two AI assistants to the user prompt displayed below.

The responses were generated based on the provided instructions and visual input from images. 

There is a provided ground truth for the instructions, but the ground truth was not given to the AI assistants when generating their responses. 

Take this context into account when making your judgment.

Steps for Evaluation:

1. Evaluate visual factuality for both responses based on the provided ground truth and visual factuality criteria.

    \quad • If the visual factuality criteria consist of **X aspects**, each aspect is worth **10/X points**.
    
    \quad • For each aspect, there may be multiple small criteria. If there are **Y small criteria in one aspect**, each small criterion is worth **10/X/Y points**.
    
2. Assign a total score out of 10 for each response.

Instructions to the AI assistants:

[INSTRUCTIONS]
\textcolor{brown}{\textless instructions\textgreater}
[END INSTRUCTIONS]

Assistant A response:

[ASSISTANT A]
\textcolor{brown}{\textless Reference Answer\textgreater}
[END ASSISTANT A]

Visual Factuality Criteria:

[VISUAL FACTUALITY CRITERIA]
\textcolor{brown}{\textless Visual Factuality Criteria\textgreater}
[END CRITERIA]

Assistant B response:

[ASSISTANT B]
\textcolor{brown}{\textless Model Prediction\textgreater}
[END ASSISTANT B]

Ground truth:

[GROUND TRUTH]
\textcolor{brown}{\textless GroundTruth\textgreater}
[END GROUND TRUTH]

Your output should evaluate visual factuality scores for each assistant and end like this:

Response A Visual Factuality Score: X/10

Response B Visual Factuality Score: Y/10

\textbf{\textcolor{blue}{Without GroundTruth: }} 

Please act as an impartial judge and evaluate the Visual Factuality of the responses provided by two AI assistants to the user prompt displayed below.

The responses were generated based on the provided instructions and visual input from images. 
Take this context into account when making your judgment.

Steps for Evaluation:

1. Evaluate visual factuality for both responses based on the visual factuality criteria.

    \quad • If the visual factuality criteria consist of **X aspects**, each aspect is worth **10/X points**.
    
    \quad • For each aspect, there may be multiple small criteria. If there are **Y small criteria in one aspect**, each small criterion is worth **10/X/Y points**.
    
2. Assign a total score out of 10 for each response.

Instructions to the AI assistants:

[INSTRUCTIONS]
\textcolor{brown}{\textless instructions\textgreater}
[END INSTRUCTIONS]

Assistant A response:

[ASSISTANT A]
\textcolor{brown}{\textless Reference Answer\textgreater}
[END ASSISTANT A]

Visual Factuality Criteria:

[VISUAL FACTUALITY CRITERIA]
\textcolor{brown}{\textless Visual Factuality Criteria\textgreater}
[END CRITERIA]

Assistant B response:

[ASSISTANT B]
\textcolor{brown}{\textless Model Prediction\textgreater}
[END ASSISTANT B]

Your output should evaluate visual factuality scores for each assistant and end like this:

Response A Visual Factuality Score: X/10

Response B Visual Factuality Score: Y/10

}
\end{AIbox} 
\caption{\textbf{Visual Factuality Judge Prompt Template of Creation-MMBench}}
\label{fig: vf_judge_creation}
\end{figure*}

%% file: main.bbl
\begin{thebibliography}{34}
\providecommand{\natexlab}[1]{#1}
\providecommand{\url}[1]{\texttt{#1}}
\expandafter\ifx\csname urlstyle\endcsname\relax
  \providecommand{\doi}[1]{doi: #1}\else
  \providecommand{\doi}{doi: \begingroup \urlstyle{rm}\Url}\fi

\bibitem[Amabile(2018)]{amabile2018creativity}
Teresa~M Amabile.
\newblock \emph{Creativity in context: Update to the social psychology of creativity}.
\newblock Routledge, 2018.

\bibitem[Bai et~al.(2023)Bai, Bai, Yang, Wang, Tan, Wang, Lin, Zhou, and Zhou]{bai2023qwen}
Jinze Bai, Shuai Bai, Shusheng Yang, Shijie Wang, Sinan Tan, Peng Wang, Junyang Lin, Chang Zhou, and Jingren Zhou.
\newblock Qwen-vl: A frontier large vision-language model with versatile abilities.
\newblock \emph{arXiv preprint arXiv:2308.12966}, 2023.

\bibitem[Chen et~al.(2024)Chen, Li, Dong, Zhang, Zang, Chen, Duan, Wang, Qiao, Lin, et~al.]{chen2024we}
Lin Chen, Jinsong Li, Xiaoyi Dong, Pan Zhang, Yuhang Zang, Zehui Chen, Haodong Duan, Jiaqi Wang, Yu Qiao, Dahua Lin, et~al.
\newblock Are we on the right way for evaluating large vision-language models?
\newblock \emph{arXiv preprint arXiv:2403.20330}, 2024.

\bibitem[Chen et~al.(2023)Chen, Wu, Wang, Su, Chen, Xing, Zhong, Zhang, Zhu, Lu, Li, Luo, Lu, Qiao, and Dai]{chen2023internvl}
Zhe Chen, Jiannan Wu, Wenhai Wang, Weijie Su, Guo Chen, Sen Xing, Muyan Zhong, Qinglong Zhang, Xizhou Zhu, Lewei Lu, Bin Li, Ping Luo, Tong Lu, Yu Qiao, and Jifeng Dai.
\newblock Internvl: Scaling up vision foundation models and aligning for generic visual-linguistic tasks.
\newblock \emph{arXiv preprint arXiv:2312.14238}, 2023.

\bibitem[Duan et~al.(2024)Duan, Yang, Qiao, Fang, Chen, Liu, Dong, Zang, Zhang, Wang, et~al.]{duan2024vlmevalkit}
Haodong Duan, Junming Yang, Yuxuan Qiao, Xinyu Fang, Lin Chen, Yuan Liu, Xiaoyi Dong, Yuhang Zang, Pan Zhang, Jiaqi Wang, et~al.
\newblock Vlmevalkit: An open-source toolkit for evaluating large multi-modality models.
\newblock In \emph{Proceedings of the 32nd ACM international conference on multimedia}, pages 11198--11201, 2024.

\bibitem[Gao et~al.(2021)Gao, Liu, Zhang, Liu, and Hao]{gao2021subcortical}
Zhenni Gao, Xiaojin Liu, Delong Zhang, Ming Liu, and Ning Hao.
\newblock Subcortical structures and visual divergent thinking: a resting-state functional mri analysis.
\newblock \emph{Brain Structure and Function}, 226\penalty0 (8):\penalty0 2617--2627, 2021.

\bibitem[Ge et~al.(2023)Ge, Chen, Chen, Chen, Chen, Chen, Xie, Yan, Zhu, Lin, et~al.]{ge2023mllm_bench}
Wentao Ge, Shunian Chen, Guiming~Hardy Chen, Junying Chen, Zhihong Chen, Nuo Chen, Wenya Xie, Shuo Yan, Chenghao Zhu, Ziyue Lin, et~al.
\newblock Mllm-bench: evaluating multimodal llms with per-sample criteria.
\newblock \emph{arXiv preprint arXiv:2311.13951}, 2023.

\bibitem[Guo et~al.(2024)Guo, Shariatmadari, Xiong, Huang, Xie, Bekiranov, and Zhang]{guo2024ideabench}
Sikun Guo, Amir~Hassan Shariatmadari, Guangzhi Xiong, Albert Huang, Eric Xie, Stefan Bekiranov, and Aidong Zhang.
\newblock Ideabench: Benchmarking large language models for research idea generation.
\newblock \emph{arXiv preprint arXiv:2411.02429}, 2024.

\bibitem[Guzik et~al.(2023)Guzik, Byrge, and Gilde]{guzik2023originality}
Erik~E Guzik, Christian Byrge, and Christian Gilde.
\newblock The originality of machines: Ai takes the torrance test.
\newblock \emph{Journal of Creativity}, 33\penalty0 (3):\penalty0 100065, 2023.

\bibitem[Hao et~al.(2025)Hao, Gu, Wang, Li, Yang, Wang, and Cheng]{hao2025can}
Yunzhuo Hao, Jiawei Gu, Huichen~Will Wang, Linjie Li, Zhengyuan Yang, Lijuan Wang, and Yu Cheng.
\newblock Can mllms reason in multimodality? emma: An enhanced multimodal reasoning benchmark.
\newblock \emph{arXiv preprint arXiv:2501.05444}, 2025.

\bibitem[Heilman(2016)]{heilman2016possible}
Kenneth~M Heilman.
\newblock Possible brain mechanisms of creativity.
\newblock \emph{Archives of Clinical Neuropsychology}, 31\penalty0 (4):\penalty0 285--296, 2016.

\bibitem[Hendrycks et~al.(2021)Hendrycks, Burns, Kadavath, Arora, Basart, Tang, Song, and Steinhardt]{hendrycks2021measuring}
Dan Hendrycks, Collin Burns, Saurav Kadavath, Akul Arora, Steven Basart, Eric Tang, Dawn Song, and Jacob Steinhardt.
\newblock Measuring mathematical problem solving with the math dataset.
\newblock \emph{arXiv preprint arXiv:2103.03874}, 2021.

\bibitem[Jia et~al.(2024)Jia, Yue, Zheng, Huang, and Lin]{jia2024simulbench}
Qi Jia, Xiang Yue, Tianyu Zheng, Jie Huang, and Bill~Yuchen Lin.
\newblock Simulbench: Evaluating language models with creative simulation tasks.
\newblock \emph{arXiv preprint arXiv:2409.07641}, 2024.

\bibitem[Liu et~al.(2023)Liu, Li, Wu, and Lee]{liu2023visual}
Haotian Liu, Chunyuan Li, Qingyang Wu, and Yong~Jae Lee.
\newblock Visual instruction tuning.
\newblock \emph{arXiv preprint arXiv:2304.08485}, 2023.

\bibitem[Liu et~al.(2024)Liu, Duan, Zhang, Li, Zhang, Zhao, Yuan, Wang, He, Liu, et~al.]{liu2024mmbench}
Yuan Liu, Haodong Duan, Yuanhan Zhang, Bo Li, Songyang Zhang, Wangbo Zhao, Yike Yuan, Jiaqi Wang, Conghui He, Ziwei Liu, et~al.
\newblock Mmbench: Is your multi-modal model an all-around player?
\newblock In \emph{European conference on computer vision}, pages 216--233. Springer, 2024.

\bibitem[Lu et~al.(2023)Lu, Bansal, Xia, Liu, Li, Hajishirzi, Cheng, Chang, Galley, and Gao]{lu2023mathvista}
Pan Lu, Hritik Bansal, Tony Xia, Jiacheng Liu, Chunyuan Li, Hannaneh Hajishirzi, Hao Cheng, Kai-Wei Chang, Michel Galley, and Jianfeng Gao.
\newblock Mathvista: Evaluating mathematical reasoning of foundation models in visual contexts.
\newblock \emph{arXiv preprint arXiv:2310.02255}, 2023.

\bibitem[Mayer(1999)]{mayer1999fifty}
Richard~E Mayer.
\newblock Fifty years of creativity research.
\newblock \emph{Handbook of creativity}, pages 449--460, 1999.

\bibitem[Mazur(2025)]{2025creativewriting}
Lech Mazur.
\newblock Llm creative story-writing benchmark.
\newblock \url{https://github.com/lechmazur/writing}, 2025.

\bibitem[Qiao et~al.(2025)Qiao, Duan, Fang, Yang, Chen, Zhang, Wang, Lin, and Chen]{qiao2025prism}
Yuxuan Qiao, Haodong Duan, Xinyu Fang, Junming Yang, Lin Chen, Songyang Zhang, Jiaqi Wang, Dahua Lin, and Kai Chen.
\newblock Prism: A framework for decoupling and assessing the capabilities of vlms.
\newblock \emph{Advances in Neural Information Processing Systems}, 37:\penalty0 111863--111898, 2025.

\bibitem[Rabeyah et~al.(2024)Rabeyah, G{\'o}es, Volpe, and Medeiros]{rabeyah2024llms}
Abdullah~Al Rabeyah, Fabr{\'\i}cio G{\'o}es, Marco Volpe, and Talles Medeiros.
\newblock Do llms agree on the creativity evaluation of alternative uses?
\newblock \emph{arXiv preprint arXiv:2411.15560}, 2024.

\bibitem[Rein et~al.(2024)Rein, Hou, Stickland, Petty, Pang, Dirani, Michael, and Bowman]{rein2024gpqa}
David Rein, Betty~Li Hou, Asa~Cooper Stickland, Jackson Petty, Richard~Yuanzhe Pang, Julien Dirani, Julian Michael, and Samuel~R Bowman.
\newblock Gpqa: A graduate-level google-proof q\&a benchmark.
\newblock In \emph{First Conference on Language Modeling}, 2024.

\bibitem[Ruan et~al.(2024)Ruan, Wang, Hong, and Sun]{ruan2024liveideabench}
Kai Ruan, Xuan Wang, Jixiang Hong, and Hao Sun.
\newblock Liveideabench: Evaluating llms' scientific creativity and idea generation with minimal context.
\newblock \emph{arXiv preprint arXiv:2412.17596}, 2024.

\bibitem[Sternberg(1997)]{sternberg1997triarchic}
Robert~J Sternberg.
\newblock The triarchic theory of intelligence.
\newblock 1997.

\bibitem[Stevenson et~al.(2022)Stevenson, Smal, Baas, Grasman, and van~der Maas]{stevenson2022putting}
Claire Stevenson, Iris Smal, Matthijs Baas, Raoul Grasman, and Han van~der Maas.
\newblock Putting gpt-3's creativity to the (alternative uses) test.
\newblock \emph{arXiv preprint arXiv:2206.08932}, 2022.

\bibitem[Wang et~al.(2024)Wang, Ma, Zhang, Ni, Chandra, Guo, Ren, Arulraj, He, Jiang, et~al.]{wang2024mmlupro}
Yubo Wang, Xueguang Ma, Ge Zhang, Yuansheng Ni, Abhranil Chandra, Shiguang Guo, Weiming Ren, Aaran Arulraj, Xuan He, Ziyan Jiang, et~al.
\newblock Mmlu-pro: A more robust and challenging multi-task language understanding benchmark.
\newblock In \emph{The Thirty-eight Conference on Neural Information Processing Systems Datasets and Benchmarks Track}, 2024.

\bibitem[Williams and G{\'o}mez-Rodr{\'\i}guez(2024)]{williams2024confederacy}
Paul Williams and Carlos G{\'o}mez-Rodr{\'\i}guez.
\newblock A confederacy of models: A comprehensive evaluation of llms on creative writing.
\newblock In \emph{UniSC Research Conference}. University of the Sunshine Coast, 2024.

\bibitem[Wu et~al.(2024)Wu, Yu, Cheng, Wang, Zhang, Xu, Ding, and Dong]{wu2024alignmmbench}
Yuhang Wu, Wenmeng Yu, Yean Cheng, Yan Wang, Xiaohan Zhang, Jiazheng Xu, Ming Ding, and Yuxiao Dong.
\newblock Alignmmbench: Evaluating chinese multimodal alignment in large vision-language models.
\newblock \emph{arXiv preprint arXiv:2406.09295}, 2024.

\bibitem[Yang et~al.(2025)Yang, Chen, Zhang, Zhao, Qian, Wang, Wang, Koripella, Movahedi, Li, et~al.]{yang2025embodiedbench}
Rui Yang, Hanyang Chen, Junyu Zhang, Mark Zhao, Cheng Qian, Kangrui Wang, Qineng Wang, Teja~Venkat Koripella, Marziyeh Movahedi, Manling Li, et~al.
\newblock Embodiedbench: Comprehensive benchmarking multi-modal large language models for vision-driven embodied agents.
\newblock \emph{arXiv preprint arXiv:2502.09560}, 2025.

\bibitem[You et~al.(2024)You, Liu, Prabhumoye, Patwary, Shoeybi, and Catanzaro]{you2024llm}
Jiaxuan You, Mingjie Liu, Shrimai Prabhumoye, Mostofa Patwary, Mohammad Shoeybi, and Bryan Catanzaro.
\newblock Llm-evolve: Evaluation for llm’s evolving capability on benchmarks.
\newblock In \emph{Proceedings of the 2024 Conference on Empirical Methods in Natural Language Processing}, pages 16937--16942, 2024.

\bibitem[Yu et~al.(2025)Yu, Shen, Meng, Lee, Yin, Cui, Xu, Zhu, Shi, Li, et~al.]{yu2025rpgbench}
Pengfei Yu, Dongming Shen, Silin Meng, Jaewon Lee, Weisu Yin, Andrea~Yaoyun Cui, Zhenlin Xu, Yi Zhu, Xingjian Shi, Mu Li, et~al.
\newblock Rpgbench: Evaluating large language models as role-playing game engines.
\newblock \emph{arXiv preprint arXiv:2502.00595}, 2025.

\bibitem[Yu et~al.(2023)Yu, Yang, Li, Wang, Lin, Liu, Wang, and Wang]{yu2023mmvet}
Weihao Yu, Zhengyuan Yang, Linjie Li, Jianfeng Wang, Kevin Lin, Zicheng Liu, Xinchao Wang, and Lijuan Wang.
\newblock Mm-vet: Evaluating large multimodal models for integrated capabilities.
\newblock \emph{arXiv preprint arXiv:2308.02490}, 2023.

\bibitem[Yue et~al.(2024)Yue, Ni, Zhang, Zheng, Liu, Zhang, Stevens, Jiang, Ren, Sun, et~al.]{yue2024mmmu}
Xiang Yue, Yuansheng Ni, Kai Zhang, Tianyu Zheng, Ruoqi Liu, Ge Zhang, Samuel Stevens, Dongfu Jiang, Weiming Ren, Yuxuan Sun, et~al.
\newblock Mmmu: A massive multi-discipline multimodal understanding and reasoning benchmark for expert agi.
\newblock In \emph{Proceedings of the IEEE/CVF Conference on Computer Vision and Pattern Recognition}, pages 9556--9567, 2024.

\bibitem[Zhang et~al.(2024)Zhang, Xu, Liu, Yu, Li, Gao, Fei, Yin, Wu, Jiang, et~al.]{zhang2024vlabench}
Shiduo Zhang, Zhe Xu, Peiju Liu, Xiaopeng Yu, Yuan Li, Qinghui Gao, Zhaoye Fei, Zhangyue Yin, Zuxuan Wu, Yu-Gang Jiang, et~al.
\newblock Vlabench: A large-scale benchmark for language-conditioned robotics manipulation with long-horizon reasoning tasks.
\newblock \emph{arXiv preprint arXiv:2412.18194}, 2024.

\bibitem[Zhang et~al.(2025)Zhang, Zhao, Fang, Li, Liu, Min, Duan, Chen, and Zhai]{zhang2025redundancyprinciplesmllmsbenchmarks}
Zicheng Zhang, Xiangyu Zhao, Xinyu Fang, Chunyi Li, Xiaohong Liu, Xiongkuo Min, Haodong Duan, Kai Chen, and Guangtao Zhai.
\newblock Redundancy principles for mllms benchmarks, 2025.

\end{thebibliography}
